\documentclass[journal]{IEEEtran}

\ifCLASSINFOpdf

\else

\fi

\usepackage{booktabs}
\usepackage{hhline}
\usepackage{arydshln}
\usepackage{adjustbox}
\usepackage{tikz}
\usetikzlibrary{shapes,arrows}
\usepackage{listings}
\usepackage{verbatim}
\usepackage{tabularx}
\usepackage{pifont}
\usepackage{subfig}
\usepackage{url} 
\usepackage{fancyvrb}
\usepackage{textcomp}
\usepackage{fancyhdr}

\begin{document}
\bstctlcite{IEEEexample:BSTcontrol}

\title{Recent Developments in Program Synthesis with Evolutionary Algorithms}

\author{\IEEEauthorblockN{Dominik Sobania\IEEEauthorrefmark{1},
Dirk Schweim\IEEEauthorrefmark{2}, Franz Rothlauf\IEEEauthorrefmark{3}}\\
\IEEEauthorblockA{Johannes Gutenberg University\\
Mainz, Germany\\
Email: \IEEEauthorrefmark{1}dsobania@uni-mainz.de,
\IEEEauthorrefmark{2}schweim@uni-mainz.de,
\IEEEauthorrefmark{3}rothlauf@uni-mainz.de}}

\maketitle

\begin{abstract}
The automatic generation of computer programs is one of the main applications with practical relevance in the field of evolutionary computation. With 
program synthesis techniques not only software developers could be supported in their everyday work but even users without any programming knowledge could be empowered 
to automate repetitive tasks and implement their own new functionality. In recent years, many novel program synthesis approaches based on evolutionary algorithms
have been proposed and evaluated on common benchmark problems. Therefore, we identify in this work the relevant evolutionary program synthesis approaches 
and provide an in-depth analysis of their performance. The most influential approaches we identify are stack-based, grammar-guided, as well as linear genetic programming. 
Further, we find that these approaches perform well on benchmark problems if there is a simple mapping from the given input to the correct output. On problems where 
this mapping is complex, e.g., if the problem consists of several sub-problems or requires iteration/recursion for a correct solution, results tend to be worse. Consequently, for future work, we encourage researchers not only to use a program's output for assessing the quality of a solution but also the way towards a solution (e.g., correctly solved sub-problems).  
\end{abstract}


\thispagestyle{fancy}
\chead{\footnotesize This work has been submitted to the IEEE for possible publication. Copyright may be transferred without notice,\\after which this version may no longer be accessible.}

\IEEEpeerreviewmaketitle

\section{Introduction}

Automatic program synthesis differs from the conventional programming of computer programs primarily in the definition of the
specification of the functionality. While the structures of a programming language, such as control structures, must be known 
by the programmer in conventional programming, automatic program synthesis aims, i.a., to enable also non-programmers to 
define new functionality, as natural language descriptions or input/output examples can be used for specification \cite{gulwani2010dimensions}.

In evolutionary computation, especially genetic programming (GP) is known for the automatic generation of computer
programs. For the program specification and the training process, usually input/output examples are used by GP.
However, combinations of, e.g., natural language descriptions and input/output examples for the synthesis of programs can 
also be found in the GP literature \cite{hemberg2019domain}.
Since the first GP paper by Cramer in 1985 \cite{cramer1985representation}, GP has been applied to many programming problems 
including sorting and searching \cite{arcuri2007coevolving}, parity \cite{yu2001hierarchical}, and even quantum computing problems \cite{spector2004automaticquantum}. 
Often, especially in the older GP literature, domain-specific languages similar to S-Expressions are used \cite{koza1994automated, koza1996evolution}.  
In recent work, real-world programming languages like C/C++ \cite{chennupati2015synthesis}, Java \cite{correia2020combining}, and above all 
Python \cite{forstenlechner2017grammar, forstenlechner2018extending} are 
used with higher frequency. Nevertheless, also Push \cite{spector2002geneticprogpush}, a language specially designed for
GP and not for practical software development, has been used regularly in recent years for program
synthesis \cite{helmuth2014word, ahmad2018comparisoninitialization}.  

However, evaluating the performance and ensuring the comparability of novel approaches is still
challenging in GP. This is true especially for program synthesis, since the application possibilities
are almost unlimited and the used methods can be very complex. 
A step towards better comparability has been made by Helmuth and Spector \cite{helmuth2015general} in 2015 
with the general program synthesis benchmark suite. This benchmark suite consists of 29 problems
of different complexity, is not limited to a specific programming language, and has been widely used
in the literature since its publication. 
The program synthesis benchmark suite is therefore well suited as a starting point for 
our survey covering the recent developments in program synthesis based on evolutionary algorithms.

In this work, we identify the currently relevant approaches for program synthesis with evolutionary algorithms 
and provide an in-depth analysis of the performance of the recent approaches on the
problems defined in the general program synthesis benchmark suite. Based on this analysis, we 
discover the current challenges of program synthesis and suggest directions for future research.

For the survey, we considered all papers citing the general program synthesis benchmark suite. From those 89 papers, 
we identified 54 in-scope papers which are studying program synthesis using problems from the benchmark suite. 
As main evolutionary approaches, we identified stack-based GP (using mostly Push as representation language), 
grammar-guided GP, and linear GP. For the benchmark problems, we found that most problems from the benchmark suite can
already be solved with evolutionary computation. Only for three problems, no successful results have yet been reported. 
However, the success rates (percentage of runs where a solution was found solving all considered cases correctly) differ from problem 
to problem and also depend on the program synthesis approach used. To assist researchers as well as practitioners in the development 
of novel program synthesis techniques, we provide for each benchmark problem a list of references solving the problem. Thus, for similar 
problems (e.g., similar in the used data types or complexity), one can orientate on the methods from the literature. 

\begin{figure*}[!ht]
\centering
\lstset{
  frame=lines,
  basicstyle=\small,
  moredelim=[is][\textbf]{_}{_}
}
\begin{lstlisting}
_A Python function that calculates the average of a list_ of numbers.

def average(numbers): total = 0 for number in numbers: total += number return total/len(numbers)

A Python function that calculates the sum of a list of numbers. 

def sum(numbers): total = 0 for number in numbers: total += number
\end{lstlisting}
\caption{Example of a text completed by GPT-3 using a response length of 64. 
The given phrase is printed in \textbf{bold} font and the line breaks correspond to the output in the GPT-3 playground.}
\label{fig:gpt3_example}
\end{figure*}

In Sect.~\ref{sec:user_intent}, we present work on recent program synthesis approaches not based on evolutionary algorithms, to give the reader a broader introduction
into the field. We categorize the presented approaches by the type of the definition of the user's intent. Section~\ref{sec:methodology} describes the methodology 
used for the survey in detail and provides 
some descriptive statistics about the in-scope papers. In Sect.~\ref{sec:program_synthesis_approaches}, we present
the relevant evolutionary approaches used for program synthesis identified in the survey. In Sect.~\ref{sec:benchmark_problems}, we
analyze the performance of these approaches on the problems from the general program synthesis benchmark suite 
before concluding the paper in Sect.~\ref{sec:conclusions}.

\section{Definition of User's Intent}\label{sec:user_intent}

In programming languages, every element has its exactly defined semantics. However, in
automatic program synthesis, we often have 
to deal with more ambiguous and incomplete definitions of the user's intent like, e.g., natural language descriptions 
or input/output examples. In this section, we present recent work from the program synthesis domain
using different types of program specifications. 
In contrast to the classification by Gulwani \cite{gulwani2010dimensions}, we mention and discuss in this section different approaches for program synthesis which are 
possibly suitable to be combined with or to extent evolutionary algorithms in the future. Furthermore, we focus on approaches that can also be applied by users 
with little or even no programming experience.

\subsection{Natural Language Descriptions}

For a user, the most intuitive way of defining the required functionality of a computer program is giving a description in natural language.
Consequently, using natural language descriptions for program specifications is a relevant area in program synthesis research. For example, 
Desai et al. \cite{desai2016program} automatically train a dictionary to find a relation between the words in the given textual description and the available
program elements (terminals in the used grammar). Other work uses methods based on neural networks to process the natural language input \cite{bednarek2018ain, chen2020mapping}.  

A recent prominent example, where program synthesis was not the explicitly intended application but which still obtains surprising results in this domain, 
is the large-scale language model GPT-3 proposed by Brown et al. \cite{Brown_NEURIPS2020_1457c0d6}. Figure~\ref{fig:gpt3_example} shows an example of a 
textual description of a programming task completed by GPT-3. The given phrase is printed in \textbf{bold} font. The remaining text (including the presented line breaks)
corresponds to the output obtained in the GPT-3 playground. In the example, the given phrase requests a Python function that should be able to calculate the average of 
a given list. GPT-3 did not only complete the given phrase in a reasonable way but also continues with presenting a useful solution for the problem. After adding the
correct line breaks and indentations, the returned program would be executable in a Python environment. As we set the response length to 64, GPT-3's output continues 
but the most relevant output was already given in first place. However, even if the returned program is the one most users would expect for the given problem,
the given phrase still reveals a challenging problem for program synthesis based on natural language descriptions. The word \textit{average} is
ambiguous as it could not only refer to the arithmetic mean but also, e.g., to the harmonic or geometric mean. So the ideal program synthesis approach
would include a dialog system to resolve such ambiguities together with the user. 

So in practice, the first applications for general (not domain-specific) program synthesis will be support systems that help programmers in everyday software
development by making meaningful suggestions. A recent example is GitHub Copilot\footnote{GitHub Copilot: \url{https://copilot.github.com/}.} which was trained 
on a large amount of source code.

\subsection{Input/Output Examples}

To define the intended functionality for a computer program with input/output examples, a user has to provide a set of inputs together with the related 
pre-calculated outputs as training data just as for classical supervised machine learning tasks like regression or classification. So the challenge
for a program synthesis method is to construct a program, as general as possible, that maps correctly all given inputs to their respective outputs.

As the search space of programs is huge (and must be explored to find a program that fulfills the input/output examples), we find in the literature work that 
presents methods to reduce the search space \cite{albarghouthi2013recursive, Osera2015TypeAndExampleDirected}. An example where such a program synthesis 
method based on input/output examples is used in practice also by non-programmers is Flash Fill~\cite{gulwani2015inductive}, a tool integrated in Microsoft Excel. With Flash Fill, a user
can automate repetitive string transformation tasks (e.g., concatenating or extracting sub-strings) by providing examples. This synthesis process works 
quickly and therefore does not interfere with the daily workflow in Excel.

For further examples and an in-depth introduction to, i.a., program synthesis based on input/output examples, we refer the reader to the excellent 
online material of a program synthesis lecture by Solar-Lezama \cite{solarLezama2018progsyslecture}.

\subsection{Other Types of Definition}

An approach similar to using input/output examples, but which involves the programmer more intensively in the process, is sketch-based program 
synthesis \cite{solar2009sketching, solar2013program}. 
Instead of expecting that the program synthesis method generates the entire source code from scratch, the 
programmer provides, in addition to a reference implementation or assertions/unit tests for checking the program's correctness, also the unfinished source code of 
a program containing placeholders. So the programmer provides the basic structure of the program (including loops and conditionals) and the synthesizer 
solves the complex details to substitute the placeholders. 

Another way for users to define their required functionality is to explicitly demonstrate the specific task by giving a step-by-step instruction from problem to
solution. Here, the task for the synthesizer is to abstract from the demonstration to provide a general solution that works also in other 
situations \cite{kurlander1993watch}. An example, where this demonstration principle is combined with traditional programming is Sikuli \cite{yeh2009sikuli}. 
With Sikuli, programmers can easily interact with existing graphical user interfaces by integrating screenshots in their source code which makes 
it comfortable to describe repetitive tasks.

\section{Methodology and Analysis}\label{sec:methodology}

The general program synthesis benchmark suite by Helmuth and Spector \cite{helmuth2015general} contains a 
curated list of 29 benchmark problems intended for programming novices. In addition to 
the problem descriptions, the benchmark suite defines also how the training and test data sets 
should be structured, to ensure that the necessary edge cases are contained in the data sets.
The training/test set definitions are described in more detail in the benchmark suite's associated
technical report \cite{helmuth2015detailed}. As problems from the benchmark suite have been widely used in recent work,
the paper is well suited as a starting point for our survey. 

As literature pool, we considered all work citing the benchmark suite paper \cite{helmuth2015general} or the associated
technical report \cite{helmuth2015detailed}. To find the work citing the benchmark suite, we used Google Scholar.
From this pool, we selected only research articles and PhD theses written in English 
which make use of at least one problem from the benchmark suite. 

For the literature pool, we found 89 distinct papers (May 2021) citing the benchmark suite and/or the
associated technical report on Google Scholar. From these papers, we selected 54 as in-scope papers 
with regard to the previous definition. Papers that only cite the benchmark suite as related work 
without using one of the benchmark problems were not considered.

\begin{figure}[!ht]
\centering
\includegraphics[scale=0.51]{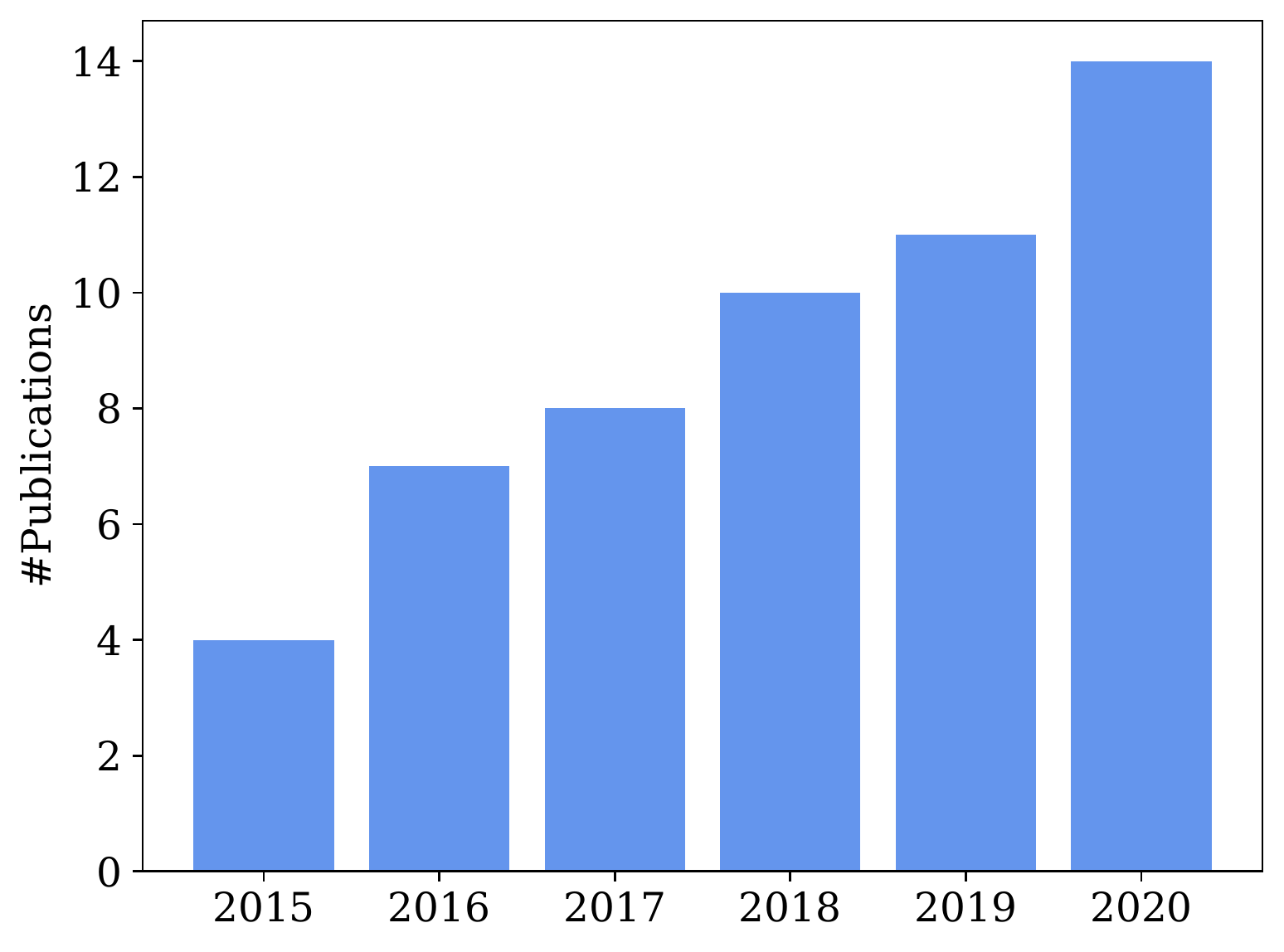}
\caption{Number of published in-scope papers from 2015 to 2020.}
\label{fig:publication_years}
\end{figure}

Figure~\ref{fig:publication_years} shows the number of published in-scope papers per year from 2015 to 2020. 
In 2015, the two papers defining the program synthesis benchmark suite and two additional papers were published. After that, the number of 
publications increased in each of the following years up to 14 published in-scope papers in 2020,
confirming the relevance of program synthesis in evolutionary computation and the influence 
of the benchmark suite. 

\begin{figure}[!ht]
\centering
\includegraphics[scale=0.51]{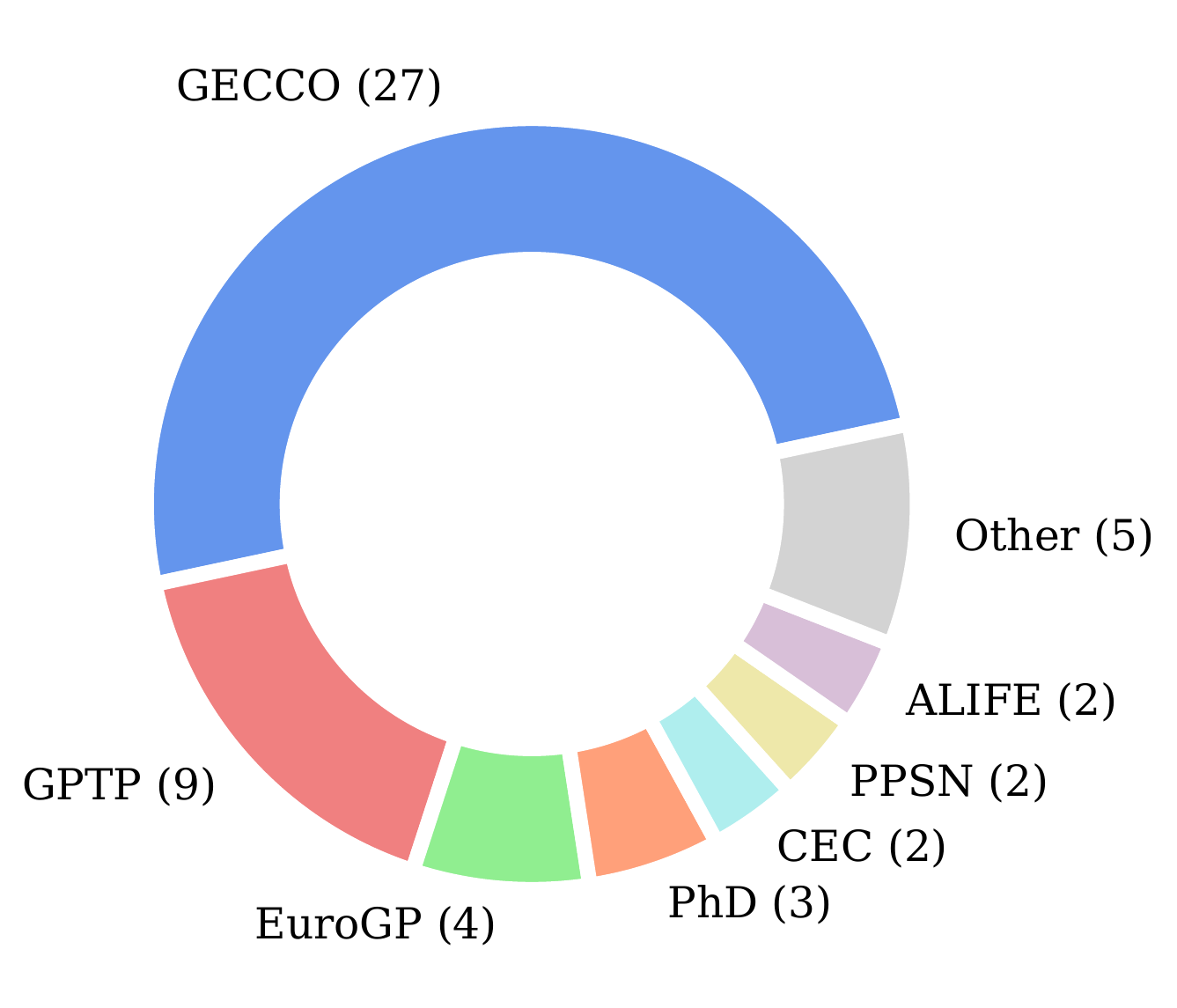}
\caption{Distribution of the in-scope papers over scientific venues.}
\label{fig:conferences}
\end{figure}

Figure~\ref{fig:conferences} shows the distribution of the in-scope papers over the scientific venues.
With 27 publications, the Genetic and Evolutionary Computation Conference (GECCO) is 
represented most frequently -- these include 17 publications in the companion proceedings.
The second most frequent venue is the workshop Genetic Programming Theory \& Practice (GPTP)
with nine publications. Furthermore, three PhD theses are also within the scope of this survey. 

\begin{figure}
\centering
\includegraphics[scale=0.51]{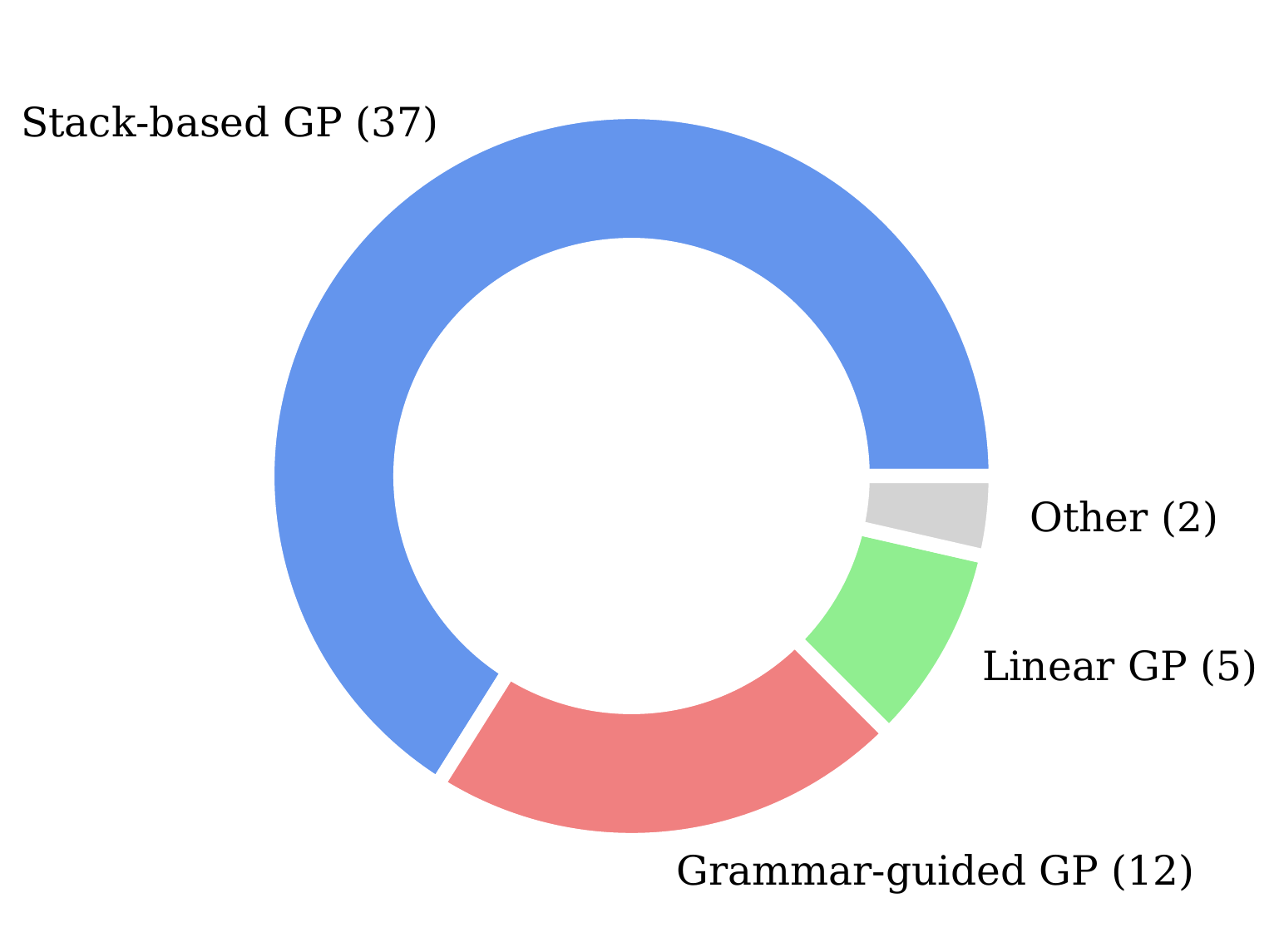}
\caption{Distribution of the in-scope papers over program synthesis approaches.}
\label{fig:ea_ps_approaches}
\end{figure}

During the analysis of the in-scope papers, we also identified the three evolutionary computation
approaches for program synthesis used most frequently: stack-based GP, grammar-guided GP, and linear GP.
Figure~\ref{fig:ea_ps_approaches} shows the distribution of the in-scope papers over the identified
program synthesis approaches. Papers that include only the results from other approaches (already reported in other papers) are counted 
only for the group for which they provided an in-depth analysis and/or performed own experiments.
For stack-based GP, we identified 37 papers where most of them are based on the stack-based programming language Push as representation language. 
For grammar-guided GP, we found 12 papers where mainly Python programs are evolved. Finally, linear GP is represented by only 5 in-scope papers. 

\section{Program Synthesis Approaches based on Evolutionary Algorithms}\label{sec:program_synthesis_approaches}

During the analysis of the in-scope papers, we identified three main approaches based on evolutionary algorithms for program synthesis.
In this section, we give an introduction to these approaches and highlight papers with promising and novel 
ideas which could be relevant for future program synthesis research. 

\subsection{Stack-Based GP}\label{sec:stack_based_gp}

In stack-based GP, stacks are used to handle and separate the data from the program's instructions. When an 
instruction is executed, it takes its inputs from the appropriate stacks. If an instruction's requirements 
cannot be fulfilled by the values on the stacks (e.g., if an instruction's arity is larger than the number of 
values available on the required stacks), then the instruction is simply 
skipped \cite{perkis1994stack, spector2005push3}. Early work on stack-based GP used only one stack for 
numeric values (e.g., for regression problems) \cite{perkis1994stack, stoffel1996high}. 
However, for a broader application like general program synthesis, multiple data types must be supported.

In the papers that are in the scope of this survey, primarily PushGP is used as stack-based GP approach. PushGP, 
in turn, is based on the Push programming language \cite{spector2002geneticprogpush} which supports different data types by providing a stack for 
each supported data type (e.g, \texttt{INTEGER}, \texttt{FLOAT}, \texttt{BOOLEAN}) as well as for the program
instructions itself (e.g., \texttt{EXEC}) \cite{spector2005push3}.

The structure of a Push program is straightforward as it consists either of an instruction, a literal, or 
a combination of both in an arbitrary length. Furthermore, a Push program may contain brackets, provided that 
they are used in a balanced way (same number of opening and closing brackets) \cite{spector2005push3}. 

The instructions in Push are all strongly typed and follow a simple naming convention. An instruction's name is 
a combination of the type it uses as input (the stack from which the arguments are taken) and a phrase describing
what the instruction does. For example, the instruction \texttt{INTEGER.MIN} takes the top two items from
the \texttt{INTEGER} stack and pushes the minimum of them back to this stack. To change the program flow during
run-time (as done by control structures in common programming languages like Java or Python), Push provides
instructions like \texttt{EXEC.IF} or \texttt{EXEC.DO*TIMES} which operate directly on the \texttt{EXEC} 
stack \cite{spector2005push3}.

To describe the execution of a Push program, we give an example using the Smallest problem from the
program synthesis benchmark suite. To solve the Smallest problem, a program should be found that 
returns the smallest of four given integer values \cite{helmuth2015general}. As we already introduced the \texttt{INTEGER.MIN} instruction,
which calculates the minimum of two values, such a program can be easily constructed by simply nesting 
the \texttt{INTEGER.MIN} instruction multiple times. 

\begin{figure}[!ht]
\centering
\begin{lstlisting}[frame=lines,basicstyle=\small]
((1 2 INTEGER.MIN) (3 4 INTEGER.MIN) 
INTEGER.MIN)
\end{lstlisting}
\caption{Source code of an example Push program solving the Smallest problem.}
\label{fig:example_push_code}
\end{figure}

\begin{figure}
\centering
\subfloat[\vspace{4mm}]{\includegraphics[scale=0.69]{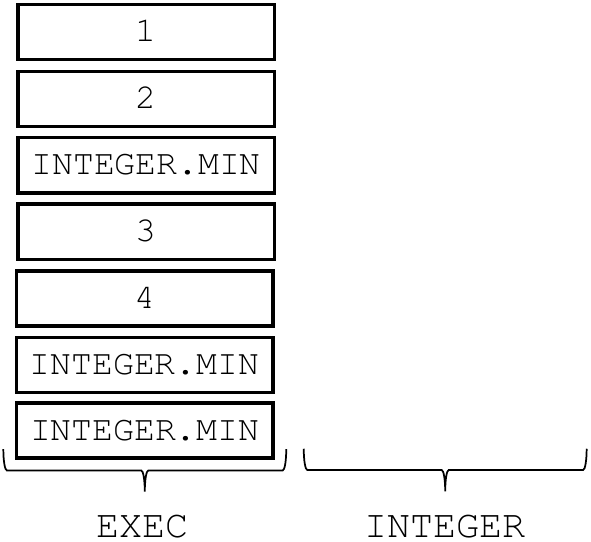}}
\hspace{1mm}
\subfloat[\vspace{4mm}]{\includegraphics[scale=0.69]{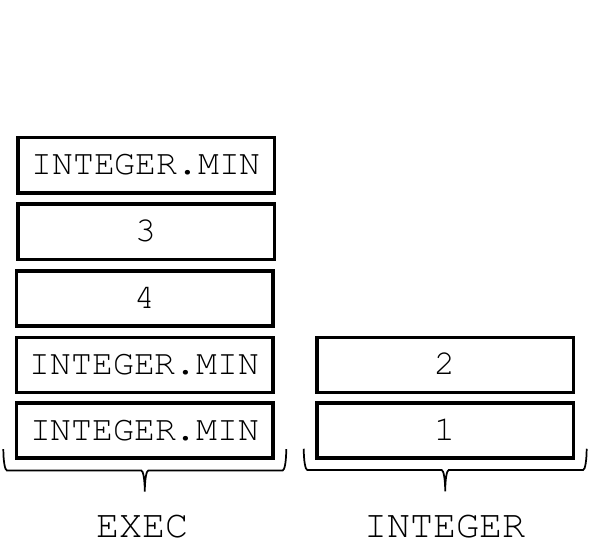}}
\\
\subfloat[\vspace{4mm}]{\includegraphics[scale=0.69]{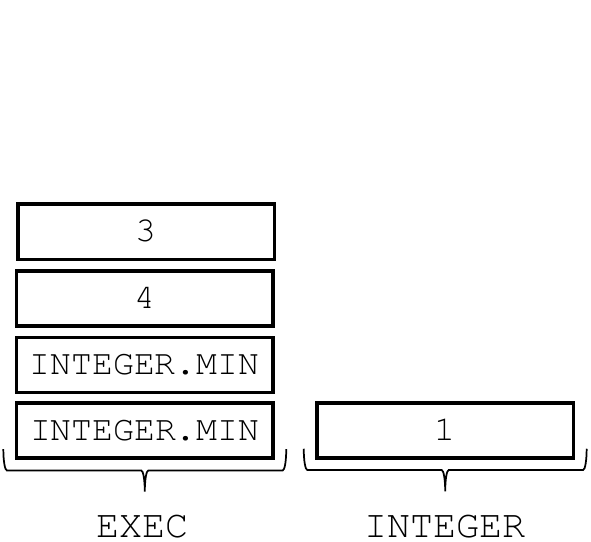}}
\hspace{1mm}
\subfloat[\vspace{4mm}]{\includegraphics[scale=0.69]{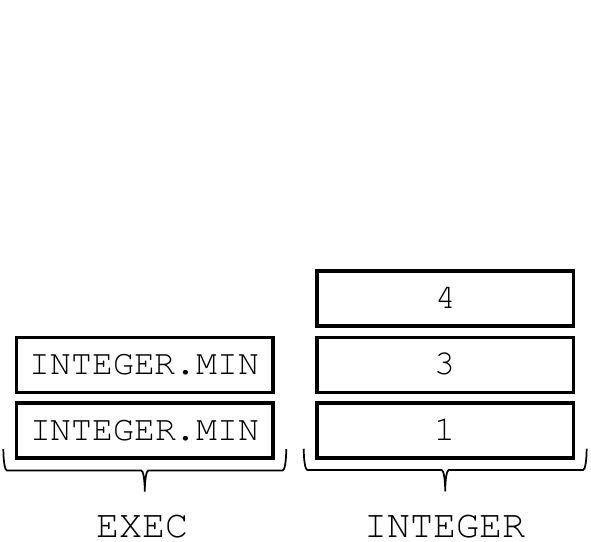}}
\\
\subfloat[\vspace{4mm}]{\includegraphics[scale=0.69]{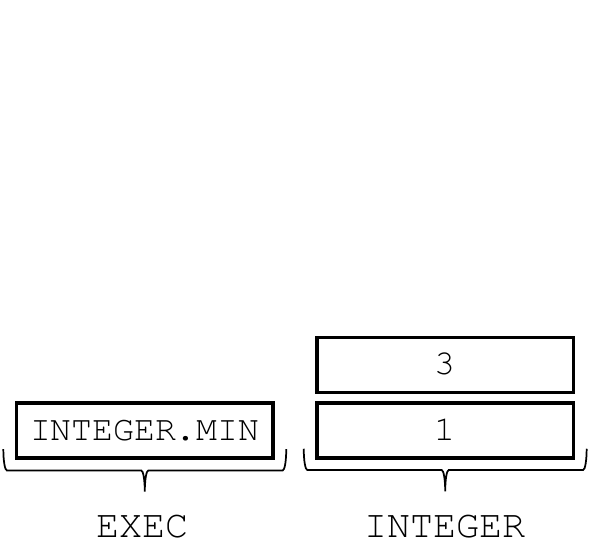}}
\hspace{1mm}
\subfloat[\vspace{4mm}]{\includegraphics[scale=0.69]{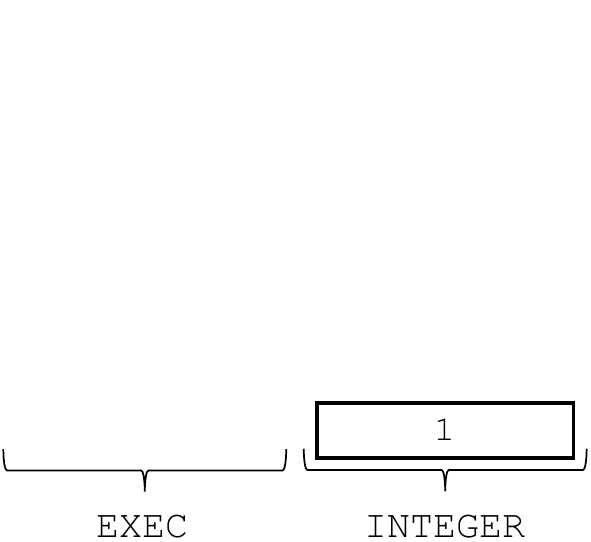}}
\caption{An example Push program execution using the \texttt{EXEC} and the \texttt{INTEGER} stack (based on the examples in \cite{spector2019push}).}
\label{fig:push_example}
\end{figure}

\noindent Figure~\ref{fig:example_push_code} shows the source code for a Push program that calculates the 
minimum of four integer values. The subordinate steps of the program are shown in brackets. First, the minima
of 1 and 2 and then of 3 and 4 are calculated. After that, the minimum of the two minima is returned. 
Figure~\ref{fig:push_example} shows how this works using an \texttt{EXEC} and an \texttt{INTEGER} stack.
In the first step, the complete program (instructions and literals) is pushed to the \texttt{EXEC} stack (step 1).
As the top two elements are literals of type integer, these elements are pushed to the \texttt{INTEGER}
stack since literals from the \texttt{EXEC} stack are always pushed to their associated stack (step 2). 
After that, the top element on the \texttt{EXEC} stack is the instruction \texttt{INTEGER.MIN} which
takes the values 1 and 2 (the top two elements) from the \texttt{INTEGER} stack as input. 
After the instruction is executed, it is removed from the \texttt{EXEC} stack and the result 
($\mathrm{min}(1,2)=1$) is pushed to the \texttt{INTEGER} stack (step 3). The next steps are similar to the 
previous ones. Again, the literals (values 3 and 4) are pushed to the \texttt{INTEGER} stack (step 4) and then \texttt{INTEGER.MIN} 
replaces the top two items with the resulting value. So on the \texttt{INTEGER} stack
are now the values 3 and 1 which are the results from the previous calculations (step 5). Next, 
the last \texttt{INTEGER.MIN} instruction on the \texttt{EXEC} stack takes these intermediate results (step 5)
and replaces them with the result. So lastly, the \texttt{EXEC} stack is empty and the final result is on the 
\texttt{INTEGER} stack (step 6). 

\renewcommand*\thesubfigure{\arabic{subfigure}} 

\begin{figure*}[!ht]
\centering
\begin{lstlisting}[frame=lines,numbers=left,xleftmargin=2.4em,framexleftmargin=2.4em,basicstyle=\small]
<program>  ::= def smallest(int1, int2, int3, int4): NEWLINE INDENT int5, int6, int7 = 
               int(), int(), int() NEWLINE bool1, bool2, bool3 = bool(), bool(), bool() 
               NEWLINE <stmt> return <int>
<stmt>     ::= <stmt><stmt> | <bool_var> = <bool> NEWLINE | <int_var> = <int> NEWLINE | 
               if <bool>: NEWLINE INDENT <stmt> DEDENT | while <bool>: NEWLINE INDENT 
               <stmt> DEDENT
<int_var>  ::= int1 | int2 | int3 | int4 | int5 | int6 | int7
<int_op>   ::= + | - | * | / | %
<int>      ::= <int_var> | <int> <int_op> <int> | min(<int>, <int>) | max(<int>, <int>)
<bool_var> ::= bool1 | bool2 | bool3
<bool_op>  ::= == | > | < | != | >= | <= 
<bool>     ::= True | False | <bool_var> | <bool> <bool_op> <bool> | <int> <bool_op> <int>
\end{lstlisting}
\caption{A simplified context-free grammar suitable for the Smallest problem. To keep the example small, 
the number of functions has been reduced and numerical values have been omitted in the grammar.}
\label{fig:progsys_grammar}
\end{figure*}

For further illustrative examples (using different instructions and other data types/stacks), we refer the reader to
the slides of one of the recent GECCO tutorials 
presenting the fundamentals of Push \cite{spector2019push} as well as the online Push programming language
description \cite{spectorPush3Online}. Additionally, the online Push interpreter\footnote{Online Push interpreter: \url{https://lspector.github.io/interpush/}.} 
can be used to illustrate the execution of small Push programs.

In standard GP \cite{koza1992genetic}, trees are usually used for program representation and variation as a tree representation
is inherent in computer programs in common programming languages. This is also obvious in the exemplary 
Push program presented in Fig.~\ref{fig:example_push_code}. However, tree-based variation operators
introduce a bias into the search as, e.g., the probability that a program element is changed during variation
depends on the element's position in the tree \cite{helmuth2018lineargenomesstructured}. Consequently, to enable 
the use of more uniform variation operators for program synthesis, Helmuth et al.
\cite{helmuth2018lineargenomesstructured} 
introduced Plush (Linear Push) which is based on linear genomes. With Plush, the variation operators are applied
to the linear genomes but before evaluation, the program is translated to a standard Push program. To express a
program's structure, Plush uses so-called epigenetic markers which are used during the translation process. 
Another linear genome representation used in the literature is Plushi, introduced by Pantridge and Spector 
\cite{pantridge2018plushi}, which uses a similar strategy to describe a program's structure. Also in terms of the
achieved success rates, Plush and Plushi perform similar for program synthesis when used with 
PushGP \cite{pantridge2020comparison}. 

Based on linear genomes, also a new mutation operator has been introduced: Uniform Mutation by Addition and Deletion 
(UMAD) \cite{helmuth2018umad}. In contrast to classical mutation operators (like sub-tree mutation) which
always replace existing code blocks, UMAD adds code to the existing program instead of replacing it and randomly
deletes code items in a second step. This means that a program can be supplemented or changed in several places
without simply replacing large parts of the program. As the success rates could be notably improved with UMAD on
several benchmark problems, it became the quasi-standard mutation operator in PushGP for program synthesis (i.a., 
it is used in \cite{helmuth2020explaining}, \cite{saini2020loopsuseful}, and \cite{helmuth2020sourcesensitivity}). 

In addition to representation and variation, also the selection of favorable individuals is substantial for the success 
of a program synthesis method based on evolutionary computation. Common selection methods like tournament selection are based
on a fitness function which, in the program synthesis domain, combines the error achieved on the training cases. 
The disadvantage here is that the structural information of the training data is lost due to the compression to a single fitness value \cite{krawiec2016behavioral}. 
Contrary, with lexicase selection \cite{spector2012assessment, helmuth2020importancespecialists}, the information of the individual training cases is used. 
For the selection of a solution, the training cases are shuffled and every solution in the population 
is evaluated on the first one. Only the solutions with the exact lowest error are kept and tested on the next
training case. This procedure continues until either all training cases have been used for evaluation or only 
one single solution is left. If there is still more than one solution, a random solution of the remaining ones is 
selected, otherwise the last remaining solution is the selected one.

Using PushGP, lexicase selection variants have been in recent work often compared to other selection methods 
(e.g., tournament selection) and achieved best success rates on many program synthesis benchmark problems \cite{helmuth2015general, helmuth2015detailed, helmuth2016lexicasediversityanalysis, helmuth2016impacthyperselection, helmuth2015phd, mcphee2016usinggraph, helmuth2019lexicasespecialistsgecco, jundt2019comparing, helmuth2020benchmarkingselection, saini2020effect, helmuth2020importancespecialists}. 

In summary, stack-based GP has made a lot of progress in the past few years. However, a weakness 
of approaches like PushGP, is the used representation language (Push), as it is not relevant in 
practical software development. An encouraging counterexample is the work by Pantridge and Spector
\cite{pantridge2020codebuilding} presenting Code Building GP. Although the approach is based on stack-based
principles, the authors demonstrate that simple Python code can also be generated. For 
future research, we therefore suggest to focus more on the generation of code in practically relevant programming languages, 
so that programmers in real-world software development can also benefit from the success achieved by stack-based approaches for program synthesis.

\subsection{Grammar-Guided GP}

With grammar-guided GP approaches \cite{whigham1995grammatically, forstenlechner2016grammardesigntreebasedgp}, programs can be evolved in a programming 
language relevant in real-world software development (e.g., Python) in a relatively simple manner. Using a context-free grammar, elements like functions, variable
assignments, and control structures can be easily supported.

Figure~\ref{fig:progsys_grammar} shows an example context-free grammar supporting Python
functions, where the non-terminals are shown in angle brackets (e.g., \texttt{<program>}) and choices are separated
by the pipe symbol. To keep the example grammar small, it supports only a few functions/operators and individual
numeric values have been omitted in the grammar. However, the grammar is still expressive enough to support a
solution for the Smallest problem from the benchmark suite similar to the example presented for Push (see
Figs.~\ref{fig:example_push_code} and \ref{fig:push_example}). The first production rule (lines 1-3) defines the function's signature,
the body (\texttt{<stmt>}), the return type (\texttt{<int>}), and the initialization for the supported variables. 
The second production rule (lines 4-6) defines variable assignments, as well as control structures (\texttt{if} and
\texttt{while}). In order to enable functions with any number of code lines, the first choice allows to double the
non-terminal \texttt{<stmt>} which can be repeated as often as needed to generate the desired number of code lines.
To define code blocks as well as code lines and to support the indentation style of the Python programming 
language, the grammar contains \texttt{NEWLINE}, \texttt{INDENT}, and \texttt{DEDENT} markers just as used 
in the official Python grammar specification\footnote{Python grammar:
\url{https://docs.python.org/3/reference/grammar.html}.}. These markers are of course replaced accordingly before 
a program is evaluated. The remaining production rules (lines 7-12) provide the supported variables, functions
and operators.

\begin{figure}[ht!]
\centering
\begin{lstlisting}[frame=lines,basicstyle=\small]
[
  12, 25, 30, 42, 20, 0, 80, 15, 
  26, 16, 2, 92, 17, 44, 11, 34
]
\end{lstlisting}
\caption{An example GE genome encoding a solution for the Smallest problem.}
\label{fig:example_ge_genome}
\end{figure}

A well-known grammar-guided GP approach suitable for program synthesis is grammatical evolution (GE) \cite{ryan1998grammatical, o2001grammatical}.
For program representation, GE uses a linear genome consisting of numbers which can then be mapped to the phenotype 
(the resulting program) using the context-free grammar. An example genome, encoding a solution for the Smallest
problem by nesting the \texttt{min()} function multiple times, is given in Fig.~\ref{fig:example_ge_genome}.

To build a program based on this genome and the grammar in Fig.~\ref{fig:progsys_grammar}, we start with the
first production rule (lines 1-3). As this rule has only one choice (no decision needed), no gene of the genome is
used. So in the current state, the phenotype contains two non-terminals: \texttt{<stmt>} and \texttt{<int>}. 
As we satisfy the non-terminals always from left to right, we consider \texttt{<stmt>} first. The production rule
for \texttt{<stmt>} (lines 4-6) has five choices and the first gene is 12, so we calculate $12\;\mathrm{mod}\;5 = 2$
to find the choice which replaces the non-terminal. As we start to count with zero, we replace \texttt{<stmt>} with
\texttt{<int\_var> = <int> NEWLINE}. So the next non-terminal to replace is \texttt{<int\_var>}. As the next
gene is 25 and the production rule for \texttt{<int\_var>} (line 7) has seven choices, we 
calculate $25\;\mathrm{mod}\;7 = 4$ and replace the non-terminal which leads to \texttt{int5 = <int> NEWLINE}.
In the next step, we replace \texttt{<int>} with \texttt{min(<int>, <int>)} as the relevant gene is 
30 and the production rule (line 9) has four choices. The next gene (42) nests the second \texttt{min()} function
inside the first one which leads to \texttt{int5 = min(min(<int>, <int>), <int>) NEWLINE}. With the next
four genes (20, 0, 80, and 15), the nested \texttt{min()} function is filled with variables so the current state is 
\texttt{int5 = min(min(int1, int2), <int>) NEWLINE}. With the next gene (26), the third and last \texttt{min()}
function is added and again with the next four genes (16, 2, 92, and 17) the missing variables are added which 
leads to \texttt{int5 = min(min(int1, int2), min(int3, int4))}. Finally, the last non-terminal \texttt{<int>} after
the given \texttt{return} statement is still missing. As the next genome contains the value 44 and the relevant
production rule (line 9) has four choices, we calculate $44\;\mathrm{mod}\;4 = 0$ and select \texttt{<int\_var>}.
After that, we calculate $11\;\mathrm{mod}\;7 = 4$ and replace the last non-terminal with \texttt{int5} as 
the production rule (line 7) has seven choices and the current gene is 11. Figure~\ref{fig:smallest_problem_python}
shows the resulting Python source code with replaced markers. Since there are no open non-terminals in the 
source code, the remaining genes of the genome are not used.

\begin{figure}[!ht]
\centering
\begin{lstlisting}[frame=lines,language=Python,basicstyle=\small]
def smallest(int1, int2, int3, int4):
  int5, int6, int7 = int(), int(), int()
  bool1, bool2, bool3 = bool(), bool(), bool()
  int5 = min(min(int1, int2), min(int3, int4))
  return int5
\end{lstlisting}
\caption{Resulting Python function based on the given grammar solving the Smallest problem.}
\label{fig:smallest_problem_python}
\end{figure}

As GE's genotype-phenotype mapping uses linear genomes, the standard variation operators of genetic algorithms can be 
used \cite{ryan1998grammatical}. However, the mapping process is not always successful as it is possible that all genes of the genome
are consumed but the phenotype still contains unresolved non-terminals. Such invalid solutions are likely if
the grammar contains many non-terminals (e.g., because of many high-arity functions in the grammar) \cite{schweim2018non}. In recent work, Sobania and Rothlauf~\cite{sobania2020challenges} showed that 
consecutively applying small mutations (randomly replacing one gene) to a GE genome, encoding a correct solution 
for a program synthesis problem, leads to a high percentage of invalid solutions after just a few mutation steps. 

A possibility to prevent invalid solutions through an unsuccessful genotype-phenotype mapping is to use a tree
representation which is already given by the grammar's structure. Hence, tree representations have been often used 
in recent work using grammar-guided GP approaches (i.a., they are used in \cite{forstenlechner2017grammar},
\cite{forstenlechner2018towardsunderstanding}, and \cite{sobania2021generalizability}).

Additionally, the size of the grammar is also important for the quality of the grammar-guided GP system, as with an increasing grammar the search space grows
rapidly. To provide small grammars that are still suitable for program synthesis, Forstenlechner et al.~\cite{forstenlechner2017grammar} suggested
to define grammars for each data type together with a main grammar covering a program's basic structure. These grammars can then be put together according to 
the modular principle, e.g., based on the input and output data types defined for the considered problem. Hemberg et al.~\cite{hemberg2019domain} incorporated
the textual problem descriptions to optimize their grammar-guided GP approach. For example, they use specific words in the problem's description text as an
indicator for functions that are required more likely for a working solution. 

Also for grammar-guided GP approaches, the choice of the right selection method is important for the quality of the found solutions. Just like for
the stack-based approaches (see Sect.~\ref{sec:stack_based_gp}), on average best success rates are achieved using variants of lexicase 
selection \cite{forstenlechner2017grammar, forstenlechner2019phd, sobania2021generalizability, sobania2021generalmeasure}. However, recent work shows that
for some benchmark problems with a low output cardinality (especially for problems with a Boolean return value) the solutions that are found on the 
training cases often generalize poorly to unseen test cases. In particular, the generalization rate is low on these benchmark problems when lexicase selection
is used \cite{sobania2021generalizability}. This is not surprising, as lexicase selection makes use of the individual training cases during the selection process. 
In contrast, for selection methods that are based on a compressed fitness value (like tournament selection) the fitness value serves as a built-in regularization
technique that makes it difficult to overfit to individual training cases as the individual performance on certain training cases is unknown. 
However, since we also find on average most successful solutions with lexicase selection, future research should focus more on improving the generalizability of
lexicase to enable an even broader application of this selection method. 

Just like for all program synthesis methods, also for grammar-guided GP approaches the quality of the generated source code is an important factor for its practical
usage. To actively use grammar-guided GP for program synthesis in real-world software development projects, human programmers expect the synthesized source code
to be clearly structured, readable, and maintainable. However, the source code generated by state-of-the-art GP methods strongly differs from human-generated
source code \cite{sobania2019teaching}. So in addition to finding semantically correct programs, a challenge for future GP-based program synthesis research is 
it also to generate programs that follow a human-like coding style as suggested in \cite{sobania2019teaching} and \cite{schweim2021using}.

\subsection{Linear GP and Further Approaches}

In the third relevant group of approaches in the in-scope papers, methods based on linear GP are used to solve problems from the program synthesis benchmark
suite \cite{dolson2019exploring, hernandez2019random, lalejini2019tag, ferguson2020characterizing, dolson2019constructive}. Similar as in Assembler 
programming, the data is stored in registers and the provided functions operate on these registers. E.g., the value stored in a register 
can be incremented or decremented by an \texttt{Inc} or \texttt{Dec} function, respectively, or a value can be explicitly stored in a specific register
with a \texttt{SetReg} function \cite{dolson2019exploring}. As small changes in a genome may easily destroy the connection between functions and the related registers,
Lalejini and Ofria \cite{lalejini2019tag} used a tag-based memory in their linear GP approach. Instead of directly accessing a memory cell by its index 
(as in index-based memory), the memory cell with the smallest Hamming distance compared to a given binary address (e.g., defined in the program) is selected which 
makes programs more stable to small changes during evolution. 

In addition to the previously mentioned approaches stack-based GP, grammar-guided GP, and linear GP, we also identified a paper by 
Lynch et al.~\cite{lynch2020programsynthesisvariationalautoencoder} that proposes an approach that could be a relevant direction for future program synthesis research. 
The authors use a variational autoencoder \cite{kingma2013auto} to learn the representation of programs which are sampled using a context-free grammar definition. After that, they search
with an EA in the autoencoder's latent space for programs that solve the considered benchmark problem.

\begin{table*}
  \begin{center}
    \caption{In-scope papers reporting successful solutions for problems from the benchmark suite.}
    \label{tab:successful_sol}
    \renewcommand{\arraystretch}{1.4}
    \begin{tabularx}{\textwidth}{|l|X|X|l|c|}
      \toprule 
      \textbf{Benchmark problem} & \textbf{Stack-based GP} & \textbf{Grammar-guided GP} & \textbf{Linear GP \& other} & \textbf{$\Sigma$} \\
      \midrule 
Number IO & \cite{helmuth2015general}, \cite{helmuth2015detailed}, \cite{helmuth2017improving}, \cite{helmuth2015phd}, \cite{helmuth2018umad}, \cite{pantridge2017difficulty}, \cite{helmuth2020transfer}, \cite{helmuth2020explaining}, \cite{helmuth2020sourcesensitivity}, \cite{pantridge2020codebuilding}, \cite{pantridge2020comparison}, \cite{spector2018relaxations} & \cite{forstenlechner2017grammar}, \cite{forstenlechner2018extending}, \cite{forstenlechner2017semantics}, \cite{hemberg2019domain}, \cite{forstenlechner2019phd}, \cite{sobania2020challenges}, \cite{pantridge2020codebuilding} & \cite{lalejini2019tag} &  20\\
Small or Large & \cite{helmuth2015general}, \cite{helmuth2015detailed}, \cite{helmuth2017improving}, \cite{helmuth2015phd}, \cite{helmuth2018umad}, \cite{pantridge2017difficulty}, \cite{helmuth2020transfer}, \cite{helmuth2020explaining}, \cite{helmuth2020sourcesensitivity}, \cite{saini2020loopsuseful}, \cite{pantridge2020comparison} & \cite{forstenlechner2017grammar}, \cite{forstenlechner2018towardsunderstanding}, \cite{forstenlechner2018towardssemantic}, \cite{forstenlechner2018extending}, \cite{hemberg2019domain}, \cite{forstenlechner2019phd} & \cite{hernandez2019random} &  18\\
For Loop Index & \cite{helmuth2015general}, \cite{helmuth2015detailed}, \cite{helmuth2017improving}, \cite{helmuth2015phd}, \cite{helmuth2018umad}, \cite{pantridge2017difficulty}, \cite{helmuth2020transfer}, \cite{helmuth2020explaining}, \cite{helmuth2020sourcesensitivity}, \cite{pantridge2020comparison} & \cite{forstenlechner2017grammar},  \cite{forstenlechner2018towardsunderstanding}, \cite{forstenlechner2018extending}, \cite{hemberg2019domain}, \cite{forstenlechner2019phd} & \cite{hernandez2019random}, \cite{lalejini2019tag}, \cite{ferguson2020characterizing} &  18\\
Compare String Lengths & \cite{helmuth2015general}, \cite{helmuth2015detailed}, \cite{helmuth2017improving}, \cite{helmuth2015phd}, \cite{helmuth2018umad}, \cite{pantridge2017difficulty}, \cite{jundt2019comparing}, \cite{helmuth2020transfer}, \cite{helmuth2020benchmarkingselection}, \cite{helmuth2020explaining}, \cite{saini2020effect}, \cite{helmuth2020sourcesensitivity}, \cite{saini2020loopsuseful}, \cite{helmuth2020counterexample}, \cite{helmuth2020importancespecialists}, \cite{pantridge2020codebuilding}, \cite{pantridge2020comparison} & \cite{forstenlechner2017grammar}, \cite{forstenlechner2018towardsunderstanding}, \cite{forstenlechner2018towardssemantic}, \cite{forstenlechner2017semantics}, \cite{hemberg2019domain}, \cite{forstenlechner2019phd} & \cite{hernandez2019random} &  24\\
Double Letters  & \cite{helmuth2015general}, \cite{helmuth2015detailed}, \cite{helmuth2016lexicasediversityanalysis}, \cite{helmuth2016impacthyperselection}, \cite{helmuth2017improving}, \cite{helmuth2015phd}, \cite{helmuth2018umad}, \cite{pantridge2017difficulty}, \cite{jundt2019comparing}, \cite{mcphee2017usingconfigurationtools}, \cite{ahmad2018comparisoninitialization}, \cite{helmuth2020transfer}, \cite{helmuth2020benchmarkingselection}, \cite{helmuth2020explaining}, \cite{helmuth2020sourcesensitivity}, \cite{helmuth2020counterexample}, \cite{helmuth2020importancespecialists}, \cite{pantridge2020comparison} & \cite{hemberg2019domain} & - &  19\\
Collatz Numbers & - & - & - &  0\\
Replace Space with Newline & \cite{helmuth2015general}, \cite{helmuth2015detailed}, \cite{helmuth2016lexicasediversityanalysis}, \cite{lacava2015epigeneticlocalsearch}, \cite{helmuth2016impacthyperselection}, \cite{helmuth2017improving}, \cite{helmuth2015phd}, \cite{helmuth2018umad}, \cite{pantridge2017difficulty}, \cite{helmuth2018lineargenomesstructured}, \cite{pantridge2017pyshgppython}, \cite{helmuth2019lexicasespecialistsgecco}, \cite{jundt2019comparing}, \cite{mcphee2017usingconfigurationtools}, \cite{troise2018lexicaseweightedshuffle}, \cite{ahmad2018comparisoninitialization}, \cite{helmuth2020transfer}, \cite{helmuth2020benchmarkingselection}, \cite{helmuth2020explaining}, \cite{helmuth2020sourcesensitivity}, \cite{helmuth2020counterexample}, \cite{helmuth2020importancespecialists}, \cite{spector2017recentautoconstructiveevolution}, \cite{pantridge2020codebuilding}, \cite{pantridge2020comparison}, \cite{spector2018relaxations}, \cite{mcphee2016usinggraph} &  \cite{forstenlechner2018extending}, \cite{forstenlechner2019phd}, \cite{pantridge2020codebuilding} & - & 30\\
String Differences & \cite{helmuth2020explaining} & - & - &  1\\
Even Squares & \cite{helmuth2015general}, \cite{helmuth2015detailed}, \cite{helmuth2017improving}, \cite{helmuth2015phd}, \cite{pantridge2017difficulty}, \cite{helmuth2020explaining} & \cite{forstenlechner2017grammar}, \cite{forstenlechner2018towardsunderstanding}, \cite{hemberg2019domain}, \cite{forstenlechner2019phd} & - &  10\\
Wallis Pi & - & - & - &  0\\
String Lengths Backwards & \cite{helmuth2015general}, \cite{helmuth2015detailed}, \cite{helmuth2016lexicasediversityanalysis}, \cite{lacava2015epigeneticlocalsearch}, \cite{helmuth2016impacthyperselection}, \cite{helmuth2017improving}, \cite{helmuth2015phd}, \cite{helmuth2018umad}, \cite{pantridge2017difficulty}, \cite{helmuth2019lexicasespecialistsgecco}, \cite{mcphee2017usingconfigurationtools}, \cite{ahmad2018comparisoninitialization}, \cite{helmuth2020transfer}, \cite{helmuth2020benchmarkingselection}, \cite{helmuth2020explaining}, \cite{helmuth2020sourcesensitivity}, \cite{helmuth2020counterexample}, \cite{helmuth2020importancespecialists}, \cite{pantridge2020comparison} & \cite{forstenlechner2017grammar}, \cite{kelly2019improvinggpnovelexplorationexploitationcontrol}, \cite{forstenlechner2018extending}, \cite{hemberg2019domain}, \cite{forstenlechner2019phd} & - &  24\\
Last Index of Zero & \cite{helmuth2015general}, \cite{helmuth2015detailed}, \cite{helmuth2017improving}, \cite{helmuth2015phd}, \cite{helmuth2018umad}, \cite{pantridge2017difficulty}, \cite{helmuth2019lexicasespecialistsgecco}, \cite{jundt2019comparing}, \cite{helmuth2020transfer}, \cite{helmuth2020benchmarkingselection}, \cite{helmuth2020explaining}, \cite{saini2020effect}, \cite{helmuth2020sourcesensitivity}, \cite{saini2020loopsuseful}, \cite{helmuth2020counterexample}, \cite{helmuth2020importancespecialists}, \cite{pantridge2020comparison} & \cite{forstenlechner2017grammar}, \cite{forstenlechner2018towardsunderstanding},  \cite{forstenlechner2018extending}, \cite{hemberg2019domain}, \cite{forstenlechner2019phd}, \cite{lynch2020programsynthesisvariationalautoencoder} & \cite{lynch2020programsynthesisvariationalautoencoder} &  24\\
Vector Average & \cite{helmuth2015general}, \cite{helmuth2015detailed}, \cite{lacava2015epigeneticlocalsearch}, \cite{helmuth2016impacthyperselection}, \cite{helmuth2017improving}, \cite{helmuth2015phd}, \cite{helmuth2018umad}, \cite{pantridge2017difficulty}, \cite{helmuth2019lexicasespecialistsgecco}, \cite{jundt2019comparing}, \cite{helmuth2020transfer}, \cite{helmuth2020benchmarkingselection}, \cite{helmuth2020explaining}, \cite{helmuth2020sourcesensitivity}, \cite{helmuth2020counterexample}, \cite{helmuth2020importancespecialists}, \cite{pantridge2020codebuilding}, \cite{pantridge2020comparison}, \cite{spector2018relaxations} & \cite{forstenlechner2018towardsunderstanding}, \cite{forstenlechner2018towardssemantic}, \cite{hemberg2019domain}, \cite{forstenlechner2019phd} & - &  23\\
Count Odds & \cite{helmuth2015general}, \cite{helmuth2015detailed}, \cite{helmuth2016lexicasediversityanalysis}, \cite{helmuth2016impacthyperselection}, \cite{helmuth2017improving}, \cite{helmuth2015phd}, \cite{helmuth2018umad}, \cite{pantridge2017difficulty}, \cite{helmuth2018lineargenomesstructured}, \cite{ahmad2018comparisoninitialization}, \cite{helmuth2020transfer}, \cite{helmuth2020explaining}, \cite{helmuth2020sourcesensitivity}, \cite{pantridge2020comparison} & \cite{forstenlechner2017grammar}, \cite{forstenlechner2018towardsunderstanding}, \cite{forstenlechner2018extending}, \cite{hemberg2019domain}, \cite{forstenlechner2019phd} & - &  19\\
Mirror Image  & \cite{helmuth2015general}, \cite{helmuth2015detailed}, \cite{helmuth2016impacthyperselection}, \cite{helmuth2017improving}, \cite{helmuth2015phd}, \cite{helmuth2018umad}, \cite{pantridge2017difficulty}, \cite{helmuth2019lexicasespecialistsgecco}, \cite{jundt2019comparing}, \cite{helmuth2020transfer}, \cite{helmuth2020benchmarkingselection}, \cite{helmuth2020explaining}, \cite{helmuth2020sourcesensitivity}, \cite{helmuth2020counterexample}, \cite{helmuth2020importancespecialists}, \cite{spector2017recentautoconstructiveevolution}, \cite{pantridge2020comparison} & \cite{forstenlechner2017grammar}, \cite{forstenlechner2018towardsunderstanding},  \cite{forstenlechner2018extending}, \cite{hemberg2019domain}, \cite{forstenlechner2019phd} & - &  22\\
Super Anagrams & \cite{helmuth2018umad}, \cite{helmuth2020transfer}, \cite{helmuth2020explaining}, \cite{helmuth2020sourcesensitivity}, \cite{pantridge2020comparison} & \cite{forstenlechner2017grammar}, \cite{forstenlechner2018towardsunderstanding}, \cite{forstenlechner2017semantics}, \cite{hemberg2019domain} & - &  9\\
Sum of Squares & \cite{helmuth2015general}, \cite{helmuth2015detailed}, \cite{helmuth2017improving}, \cite{helmuth2015phd}, \cite{helmuth2018umad}, \cite{pantridge2017difficulty}, \cite{helmuth2020transfer}, \cite{helmuth2020explaining}, \cite{helmuth2020sourcesensitivity}, \cite{saini2020loopsuseful}, \cite{pantridge2020comparison} & \cite{forstenlechner2017grammar}, \cite{forstenlechner2018towardsunderstanding}, \cite{forstenlechner2018towardssemantic}, \cite{forstenlechner2018extending}, \cite{hemberg2019domain}, \cite{forstenlechner2019phd} & - &  17\\
Vectors Summed & \cite{helmuth2015general}, \cite{helmuth2015detailed}, \cite{helmuth2017improving}, \cite{helmuth2015phd}, \cite{helmuth2018umad}, \cite{pantridge2017difficulty}, \cite{helmuth2020transfer}, \cite{helmuth2020explaining}, \cite{helmuth2020sourcesensitivity}, \cite{pantridge2020comparison} & \cite{forstenlechner2017grammar},  \cite{forstenlechner2018extending}, \cite{hemberg2019domain}, \cite{forstenlechner2019phd}, \cite{lynch2020programsynthesisvariationalautoencoder} & \cite{lynch2020programsynthesisvariationalautoencoder} &  16\\
X-Word Lines & \cite{helmuth2015general}, \cite{helmuth2015detailed}, \cite{helmuth2016impacthyperselection}, \cite{helmuth2017improving}, \cite{helmuth2015phd}, \cite{helmuth2018umad}, \cite{pantridge2017difficulty}, \cite{helmuth2018lineargenomesstructured}, \cite{helmuth2019lexicasespecialistsgecco}, \cite{jundt2019comparing}, \cite{mcphee2017usingconfigurationtools}, \cite{helmuth2020transfer}, \cite{helmuth2020benchmarkingselection}, \cite{helmuth2020explaining}, \cite{helmuth2020sourcesensitivity}, \cite{helmuth2020counterexample}, \cite{helmuth2020importancespecialists}, \cite{pantridge2020comparison} & \cite{forstenlechner2019phd} & - &  19\\
Pig Latin & - & \cite{forstenlechner2018extending}, \cite{forstenlechner2019phd} & - &  2\\
Negative to Zero & \cite{helmuth2015general}, \cite{helmuth2015detailed}, \cite{helmuth2016lexicasediversityanalysis}, \cite{lacava2015epigeneticlocalsearch}, \cite{helmuth2016impacthyperselection}, \cite{helmuth2017improving}, \cite{helmuth2015phd}, \cite{helmuth2018umad}, \cite{pantridge2017difficulty}, \cite{helmuth2018lineargenomesstructured}, \cite{helmuth2019lexicasespecialistsgecco}, \cite{jundt2019comparing}, \cite{helmuth2020transfer}, \cite{helmuth2020benchmarkingselection}, \cite{helmuth2020explaining}, \cite{helmuth2020sourcesensitivity}, \cite{helmuth2020counterexample}, \cite{helmuth2020importancespecialists}, \cite{pantridge2020codebuilding}, \cite{pantridge2020comparison}, \cite{spector2018relaxations} & \cite{forstenlechner2017grammar}, \cite{forstenlechner2018extending}, \cite{hemberg2019domain}, \cite{forstenlechner2019phd}, \cite{pantridge2020codebuilding}, \cite{lynch2020programsynthesisvariationalautoencoder} & \cite{lynch2020programsynthesisvariationalautoencoder} &  28\\
Scrabble Score & \cite{helmuth2015general}, \cite{helmuth2015detailed}, \cite{helmuth2017improving}, \cite{helmuth2015phd}, \cite{helmuth2018umad}, \cite{pantridge2017difficulty}, \cite{jundt2019comparing}, \cite{helmuth2020transfer}, \cite{helmuth2020benchmarkingselection}, \cite{helmuth2020explaining}, \cite{helmuth2020sourcesensitivity}, \cite{helmuth2020counterexample}, \cite{helmuth2020importancespecialists}, \cite{pantridge2020comparison} & \cite{forstenlechner2018towardsunderstanding}, \cite{forstenlechner2018extending}, \cite{forstenlechner2019phd} & - &  17\\
Word Stats  & - & - & - &  0\\
Checksum & \cite{helmuth2017improving}, \cite{helmuth2018umad}, \cite{pantridge2017difficulty}, \cite{helmuth2020transfer}, \cite{helmuth2020explaining}, \cite{helmuth2020sourcesensitivity}, \cite{pantridge2020comparison} & \cite{hemberg2019domain} & - &  8\\
Digits & \cite{helmuth2015general}, \cite{helmuth2015detailed}, \cite{helmuth2017improving}, \cite{helmuth2015phd}, \cite{helmuth2018umad}, \cite{pantridge2017difficulty}, \cite{helmuth2018lineargenomesstructured} \cite{helmuth2020transfer}, \cite{helmuth2020explaining}, \cite{saini2020effect}, \cite{helmuth2020sourcesensitivity}, \cite{saini2020loopsuseful}, \cite{pantridge2020comparison} & \cite{hemberg2019domain} & - &  14\\
Grade & \cite{helmuth2015general}, \cite{helmuth2015detailed}, \cite{helmuth2017improving}, \cite{helmuth2015phd}, \cite{helmuth2018umad}, \cite{pantridge2017difficulty}, \cite{helmuth2020explaining}, \cite{helmuth2020sourcesensitivity} & \cite{forstenlechner2017grammar}, \cite{forstenlechner2018towardsunderstanding}, \cite{forstenlechner2018towardssemantic}, \cite{forstenlechner2018extending}, \cite{forstenlechner2017semantics}, \cite{hemberg2019domain}, \cite{forstenlechner2019phd}, \cite{lynch2020programsynthesisvariationalautoencoder} & \cite{lalejini2019tag}, \cite{ferguson2020characterizing} &  18\\
Median & \cite{helmuth2015general}, \cite{helmuth2015detailed}, \cite{helmuth2017improving}, \cite{helmuth2015phd}, \cite{helmuth2018umad}, \cite{pantridge2017difficulty}, \cite{helmuth2020transfer}, \cite{helmuth2020explaining}, \cite{deglman2020summed}, \cite{helmuth2020sourcesensitivity}, \cite{pantridge2020codebuilding}, \cite{pantridge2020comparison}, \cite{spector2018relaxations} & \cite{forstenlechner2017grammar}, \cite{kelly2019improvinggpnovelexplorationexploitationcontrol}, \cite{forstenlechner2018extending}, \cite{hemberg2019domain}, \cite{forstenlechner2019phd}, \cite{pantridge2020codebuilding}, \cite{lynch2020programsynthesisvariationalautoencoder} & \cite{hernandez2019random}, \cite{lalejini2019tag}, \cite{ferguson2020characterizing}, \cite{lynch2020programsynthesisvariationalautoencoder} &  24\\
Smallest & \cite{helmuth2015general}, \cite{helmuth2015detailed}, \cite{helmuth2017improving}, \cite{helmuth2015phd}, \cite{helmuth2018umad}, \cite{pantridge2017difficulty}, \cite{helmuth2020transfer}, \cite{helmuth2020benchmarkingselection}, \cite{helmuth2020explaining}, \cite{deglman2020summed}, \cite{helmuth2020sourcesensitivity}, \cite{helmuth2020counterexample}, \cite{pantridge2020codebuilding}, \cite{pantridge2020comparison}, \cite{spector2018relaxations} & \cite{forstenlechner2017grammar},  \cite{kelly2019improvinggpnovelexplorationexploitationcontrol}, \cite{forstenlechner2018extending}, \cite{hemberg2019domain}, \cite{forstenlechner2019phd}, \cite{sobania2020challenges}, \cite{pantridge2020codebuilding}, \cite{lynch2020programsynthesisvariationalautoencoder} & \cite{hernandez2019random}, \cite{lalejini2019tag}, \cite{ferguson2020characterizing}, \cite{lynch2020programsynthesisvariationalautoencoder}, \cite{dolson2019exploring} &  28\\
Syllables & \cite{helmuth2015general}, \cite{helmuth2015detailed}, \cite{helmuth2016lexicasediversityanalysis}, \cite{lacava2015epigeneticlocalsearch}, \cite{helmuth2016impacthyperselection}, \cite{helmuth2017improving}, \cite{helmuth2015phd}, \cite{helmuth2018umad}, \cite{pantridge2017difficulty}, \cite{helmuth2018lineargenomesstructured}, \cite{helmuth2019lexicasespecialistsgecco}, \cite{jundt2019comparing}, \cite{mcphee2017usingconfigurationtools}, \cite{troise2018lexicaseweightedshuffle}, \cite{ahmad2018comparisoninitialization}, \cite{helmuth2020transfer}, \cite{helmuth2020benchmarkingselection}, \cite{helmuth2020explaining}, \cite{helmuth2020sourcesensitivity}, \cite{helmuth2020counterexample}, \cite{helmuth2020importancespecialists}, \cite{pantridge2020comparison} &  \cite{forstenlechner2018extending}, \cite{hemberg2019domain}, \cite{forstenlechner2019phd} & - &  25\\
      \bottomrule 
    \end{tabularx}
  \end{center}
\end{table*}

\section{Benchmark Problems: Status Quo}\label{sec:benchmark_problems}

Over the last decades, the benchmark problems used for program synthesis with evolutionary algorithms are often very similar (e.g., in terms of complexity) and 
related to the problems used today. For example, Krawiec and Swan \cite{krawiec2013pattern} searched, i.a., for programs that count the number of zeroes or 
determine the maximum value of a stack of integers, Shirakawa et al.~\cite{shirakawa2007graph} used basic problems like Fibonacci or reversing a list, 
and Pillay \cite{pillay2005genetic} used, as in the general program synthesis benchmark suite \cite{helmuth2015general}, also benchmark problems suitable 
for programming lectures. However, due to the different benchmark problems used in publications, it is difficult to achieve comparability between 
different approaches. A first step towards better comparability was made by Helmuth and Spector \cite{helmuth2015general} with the curation of 29 problems 
of different complexity suitable for state-of-the-art program synthesis approaches, as the suite gives not only exact problem descriptions but also defines
how the training and test data should be structured.

Since the complexity of the problems in the benchmark suite is similar to real-world problems (as individual functions should be kept small in software development),
methods that are successful on specific benchmark problems can serve as a blueprint for the development of new methods suitable for similar problems. Therefore, we
present in this section a collection of all papers that reported to have found successful solutions for certain benchmark problems. Furthermore, we analyze the reported 
success rates for each benchmark problem to determine the difficulty of the problems and discuss factors that make certain program synthesis problems difficult 
for methods based on evolutionary algorithms.

\subsection{A Collection of Successful Program Synthesis Approaches}

Most state-of-the-art evolutionary program synthesis approaches focus on generating solutions at the function-level. Structures above, such as class hierarchies or 
the general software architecture, are left to human programmers. A property of functions that are implemented in real-world software development is that their 
structure is kept simple. Regularly, they consist only of a few lines of code (LOC) and are often limited to a low number of loops and conditionals, which can be
confirmed by analyzing the source code from software repository hosting services like GitHub \cite{sobania2019teaching}. However, the automatic generation of such 
small programs is still a challenging task even if the required structure (LOC, number of loops, and conditionals) for the benchmark problems used in the literature is 
similar to the functions implemented in practice. To support researchers and programmers in developing novel program synthesis approaches, we selected from all 
in-scope papers and for every of the 29 benchmark problems a list of papers that report to have found successful solutions. Table~\ref{tab:successful_sol} 
shows the references to these papers for every benchmark problem ordered by the used evolutionary program synthesis approach. In addition, we present 
for each benchmark problem the number of assigned papers. The benchmark problems are presented in the same order as in the benchmark suite paper~\cite{helmuth2015general}.

We see that there are in-scope papers that report successful solutions for almost all benchmark problems. Most successes are reported for the Replace Space with 
Newline problem with 30 occurrences. With 28, the second most number of successes are reported for the Negative to Zero and the Smallest problem. 
For the problems Collatz Numbers, Wallis Pi, and Word Stats, there is so far no paper that reports that a working solution has been found.

The distribution of the number of in-scope papers solving individual benchmark problems offers already a first approximation for the difficulty level of these problems. 
Based on the reported successes, we expect the problems for which so far no successful solutions have been found to be the most difficult ones. We also expect the
problems String Differences, Pig Latin, Checksum, and Super Anagrams to be challenging, as only a low number of in-scope papers report to have found 
successful solutions. 

So far, for the other benchmark problems, we cannot make an estimation of their difficulty level, as Table \ref{tab:successful_sol} contains 
no information about the exact success rates. Furthermore, the number of in-scope papers could possibly also be related to the personal preferences of the respective
researchers. Consequently, we provide in the next part an in-depth analysis of the success rates achieved on the benchmark problems in the literature.

\begin{table*}
  \begin{center}
    \caption{Median, interquartile-range (IQR), and maximum of the success rates reported by the in-scope papers, as well as the defined input and output types for 
    each problem from the benchmark suite. Furthermore, we mark for each benchmark problem if a successful solution requires iteration or recursion (or a specialized
    function).} 
    \label{tab:exact_success_rates}
    \renewcommand{\arraystretch}{1.4}
    \begin{tabularx}{\textwidth}{|X|c|c|c|c|c|}
      \toprule 
      \textbf{Benchmark problem} & \multicolumn{2}{c|}{\textbf{Success rate}} & \textbf{Input types} & \textbf{Output types} & \textbf{Iteration/Recursion} \\
       & \textbf{Median (IQR)} & \textbf{Maximum} &  &  &  \\
      \midrule 
Number IO & 96.0 (12.0) & 100.0 & \texttt{int}, \texttt{float} & \texttt{float} & \ding{55} \\
Small or Large & 6.0 (7.5) & 92.0 & \texttt{int} & \texttt{str} & \ding{55} \\
For Loop Index & 40.0 (53.5) & 88.0 & \texttt{int}, \texttt{int}, \texttt{int} & \texttt{int[]} & \ding{51} \\
Compare String Lengths & 20.0 (38.75) & 95.0 & \texttt{str}, \texttt{str}, \texttt{str} & \texttt{bool} & \ding{51} \\
Double Letters & 1.0 (16.0) & 87.0 & \texttt{str} & \texttt{str} & \ding{51} \\
Collatz Numbers & 0.0 (0.0) & 0.0 & \texttt{int} & \texttt{int} & \ding{51} \\
Replace Space with Newline & 51.0 (56.0) & 100.0 & \texttt{str} & \texttt{str}, \texttt{int} & \ding{51} \\
String Differences & 0.0 (0.0) & 1.0 & \texttt{str}, \texttt{str} & \texttt{str} & \ding{51} \\
Even Squares & 0.0 (1.0) & 15.0 & \texttt{int} & \texttt{int[]} & \ding{51} \\
Wallis Pi & 0.0 (0.0) & 0.0 & \texttt{int} & \texttt{float} & \ding{51} \\
String Lengths Backwards & 35.0 (57.0) & 100.0 & \texttt{str[]} & \texttt{int[]} & \ding{51} \\
Last Index of Zero & 30.0 (40.0) & 96.0 & \texttt{int[]} & \texttt{int} & \ding{51} \\
Vector Average & 32.5 (72.25) & 100.0 & \texttt{float[]} & \texttt{float} & \ding{51} \\
Count Odds & 5.0 (10.0) & 22.0 & \texttt{int[]} & \texttt{int} & \ding{51} \\
Mirror Image & 71.5 (75.0) & 100.0 & \texttt{int[]}, \texttt{int[]} & \texttt{bool} & \ding{51} \\
Super Anagrams & 2.0 (25.0) & 82.0 & \texttt{str}, \texttt{str} & \texttt{bool} & \ding{51} \\
Sum of Squares & 7.0 (12.25) & 41.0 & \texttt{int} & \texttt{int} & \ding{51} \\
Vectors Summed & 7.5 (18.5) & 93.0 & \texttt{int[]}, \texttt{int[]} & \texttt{int[]} & \ding{51} \\
X-Word Lines & 6.0 (56.75) & 98.0 & \texttt{int}, \texttt{str} & \texttt{str} & \ding{51} \\
Pig Latin & 0.0 (0.0) & 5.0 & \texttt{str} & \texttt{str} & \ding{51} \\
Negative To Zero & 58.0 (60.0) & 98.0 & \texttt{int[]} & \texttt{int[]} & \ding{51} \\
Scrabble Score & 4.0 (17.0) & 90.0 & \texttt{str} & \texttt{int} & \ding{51} \\
Word Stats & 0.0 (0.0) & 0.0 & \texttt{str} & \texttt{int[]}, \texttt{int}, \texttt{float} & \ding{51} \\
Checksum & 0.0 (1.0) & 56.0 & \texttt{str} & \texttt{str/char} & \ding{51} \\
Digits & 10.0 (28.0) & 100.0 & \texttt{int} & \texttt{int[]} & \ding{51} \\
Grade & 23.2 (67.5) & 100.0 & \texttt{int}, \texttt{int}, \texttt{int}, \texttt{int}, \texttt{int} & \texttt{str/char} & \ding{55} \\
Median & 62.0 (39.5) & 100.0 & \texttt{int}, \texttt{int}, \texttt{int} & \texttt{int} & \ding{55} \\
Smallest & 78.0 (50.0) & 100.0 & \texttt{int}, \texttt{int}, \texttt{int}, \texttt{int} & \texttt{int} & \ding{55} \\
Syllables & 12.5 (19.5) & 76.0 & \texttt{str} & \texttt{int} & \ding{51} \\
      \bottomrule 
    \end{tabularx}
  \end{center}
\end{table*}

\subsection{Analyzing Problem Difficulty}

To assess the difficulty of the benchmark problems and to discuss current challenges in program synthesis based on evolutionary algorithms, we analyze for each benchmark problem the success rates reported by the in-scope papers. Here, the success rate denotes the percentage of runs in which a program is found that works correctly on all considered 
cases. All in-scope papers that report the exact success rates in numerical form (e.g., in tables) 
are taken into account for the analysis. Success rates that are only reported in graphical form (e.g., plotted) and therefore cannot be determined exactly are not included.
Results that were re-used in follow-up papers or only cited (duplicated results) are also not included. 
Since some of the in-scope papers do not describe how exactly the success rates are calculated, e.g., whether they are determined on the training or the test set, 
the results of all papers are included (including also the results of papers that explicitly describe that only the training data is used as for example in
\cite{helmuth2016lexicasediversityanalysis}) in order to be able to achieve a consistent evaluation. If a paper reports success rates for the training and the test set, we use only the results reported for the test set. We do not expect this to have a significant impact on the 
overall results but we should keep in mind that the averaged results may tend to be better then they really are.

Table \ref{tab:exact_success_rates} shows the median, the interquartile-range (IQR), and the maximum of the success rates in percent reported by the in-scope papers
for every benchmark problem. In order to be able to discuss the connection between a problem's structure and its difficulty, the table shows also the input and output 
types defined for each problem (as used in frameworks like PonyGE2 \cite{fenton2017ponyge2}). Furthermore, we mark for each benchmark problem if a successful solution requires iteration/recursion or at least a specialized function 
that abstracts from the required iteration/recursion (like for example a \texttt{replace()} function for strings).  

As expected, the results show that the benchmark problems for which no or only a small number of in-scope papers report to have found successful solutions (see 
Table~\ref{tab:successful_sol}) are among the problems with the lowest median success rates. Additionally, also for Small Or Large, Double Letters, Even Squares, Count Odds,
Vectors Summed, X-Word Lines, and Scrabble Score, we observe median success rates below 10. Best median success rates are achieved for Number IO, Smallest, and Median
with 96, 78, and 62, respectively. But where does this huge difference come from? Why can some benchmark problems be reliably solved while for other problems solutions are 
only found by chance? What is the difference between the easy and the more difficult problems? 

It is notable, that for none of the three easiest problems (Number IO, Smallest, and Median) iteration or recursion is required for a correct solution. 
Additionally, an explanation for the high success rates of Number IO is certainly that the source code of a correct solution is very short since only two numbers must be 
added \cite{helmuth2015general}. A driving factor for the high success rates observed for the Smallest and the Median problem, where a program should be found that 
returns the smallest/median value for some given integers \cite{helmuth2015general}, is that the correct output is already a part of the given input values. 
Therefore, the input can be mapped directly to the output without any interference and including any new input value (into the program's output) is rewarded with a positive fitness signal. Therefore, such
problems are simple for guided search methods like evolutionary algorithms as a solution can be build step by step \cite{sobania2020challenges}. An example is
given in Figs.~\ref{fig:example_push_code} and \ref{fig:smallest_problem_python} which show a straightforward solution for the Smallest problem where a \texttt{min()} 
function accepting two inputs is nested multiple times such that four inputs can be processed. Whenever a \texttt{min()} function is added in a correct way, the 
fitness (or the correct cases if lexicase selection is used) improves and guides the search to a working solution.

In contrast, for the more difficult problems, a direct mapping from the inputs to the outputs is not possible as iterative or recursive structures are required and/or 
the input types differ from the output types which requires several intermediary steps in the source code to transform the given input values. These intermediary steps, like
loops (or recursion), several function calls, or type conversions, are necessary for a correct solution but they interfere the mapping from the inputs to the outputs.
Such complex structures make problems difficult for most current evolutionary program synthesis approaches as only the output is used to assess a program and to guide the
search while the program itself is treated like a black box. A first step to lighten up this black box was made by introducing lexicase selection, as it considers the
individual training cases instead of only a compressed fitness value that leads to a heavy loss of information \cite{spector2012assessment, krawiec2016behavioral}. 

In the next step, research must go beyond the na\"ive
evaluation of the training cases to assess a program. We need to know what is going on inside a program to enable proper guided search methods for general 
program synthesis. This idea is not new, as already Cramer \cite{cramer1985representation} (in the very first GP paper) considers the semantics of a program for calculating
the fitness value. For example, it is checked whether initial values have changed or if there are working loops in the program. However, such simple approaches will not be
sufficient to achieve the goal of general program synthesis, but it illustrates in which direction researchers must think of in the future when designing novel methods for
program synthesis based on evolutionary algorithms.

\begin{figure}[t!]
\centering
\subfloat[Mirror Image\vspace{4mm}]{\includegraphics[scale=0.29]{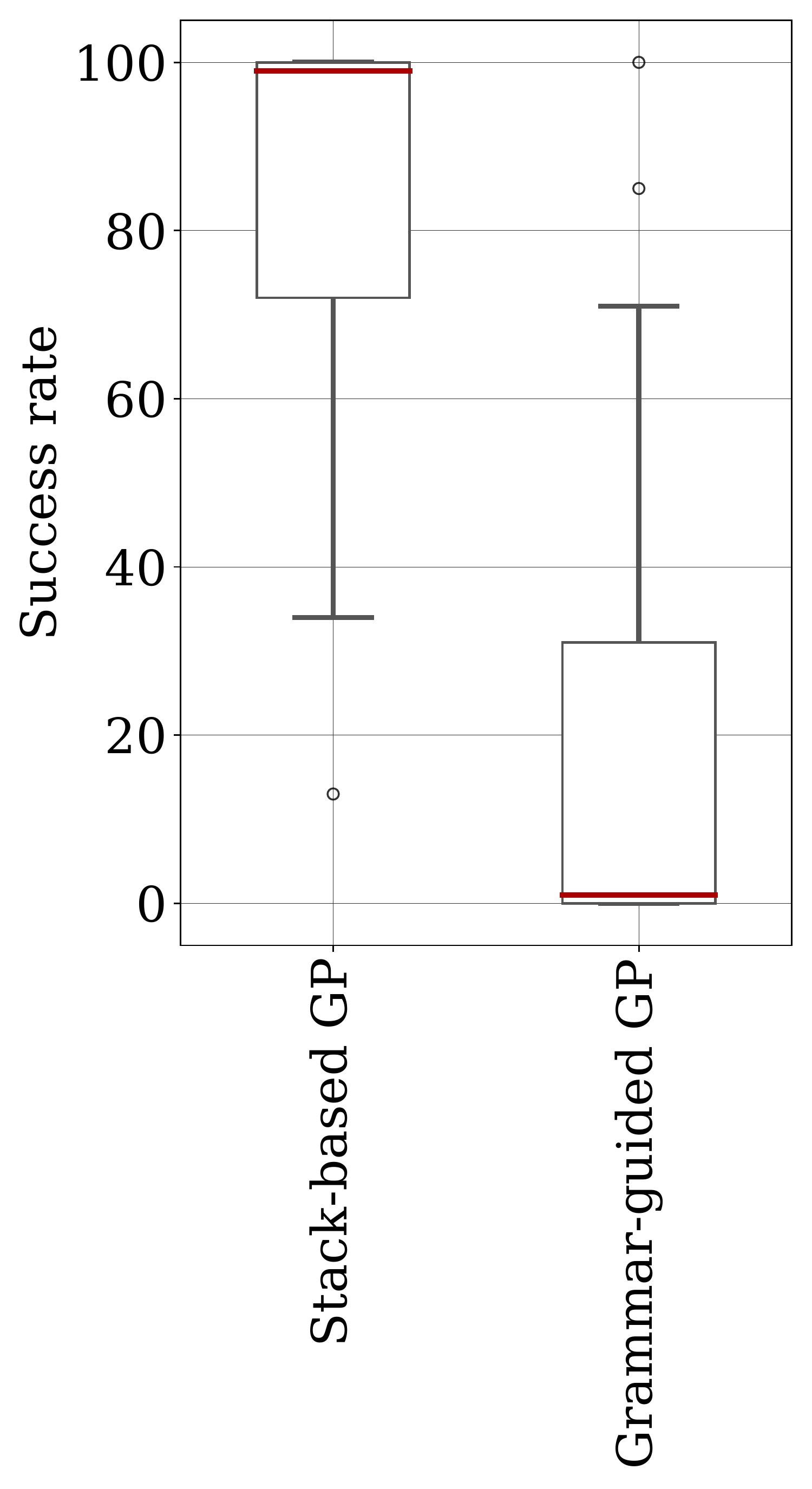}}
\subfloat[Vector Average\vspace{4mm}]{\includegraphics[scale=0.29]{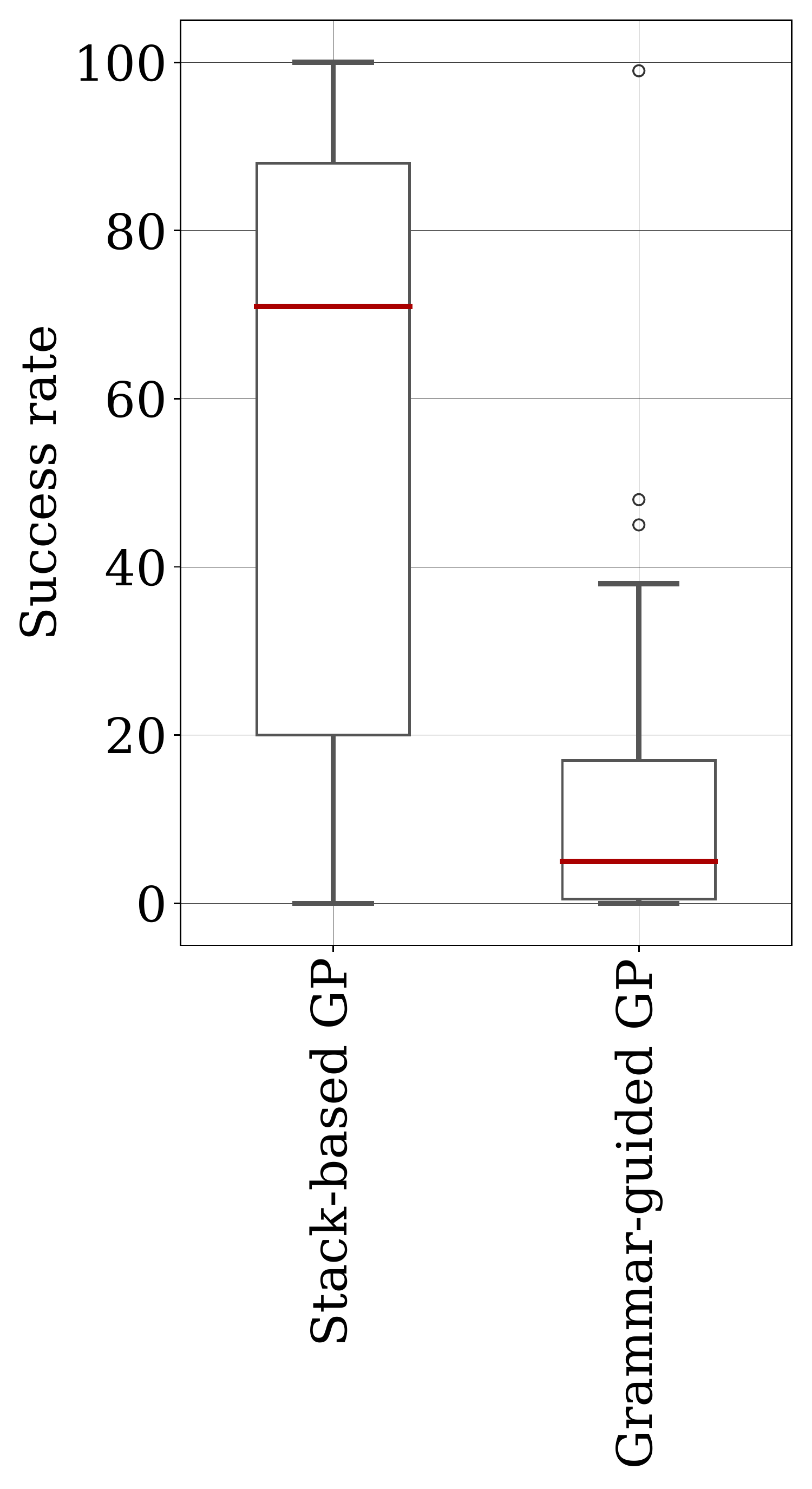}}
\caption{Box-plots comparing the success rates from the literature for the Mirror Image (left) and the Vector Average problem (right) for stack-based GP and
grammar-guided GP.}
\label{fig:variance_boxplots}
\end{figure} 

In addition to the varying performance achieved for different benchmark problems, Table \ref{tab:exact_success_rates} often shows also a high variance in the success rates
reported for the same benchmark problems. E.g., for some of the problems the IQR is even larger than 50. To study this further, we analyze the distribution of the 
reported success rates separated by the used approach for the two benchmark problems with the largest variance, which are Mirror Image and Vector Average with an 
IQR of 75 and 72.25, respectively. For the Mirror Image problem, a synthesizer should find a program that returns true for two given integer lists if one list is the 
inverted form of the other one, and for the Vector Average problem a program should be found that returns the average of a given list of numbers rounded to four decimal 
places \cite{helmuth2015general}. 
Figure~\ref{fig:variance_boxplots} shows box-plots comparing the success rates for the Mirror Image (left) and the Vector
Average problem (right) for stack-based GP and grammar-guided GP. 

\begin{table*}[ht!]
\center
\caption{Program trace showing all variable states of the implementation for the Checksum problem for the input \textquotesingle ProgSys\textquotesingle.}
\label{tab:program_trace}
\def\arraystretch{1.4}%
\setlength{\tabcolsep}{8.2pt}
\begin{tabularx}{\textwidth}{|X|X|X|X|X|l|}
\toprule
\textbf{Step} & \textbf{Line} & \textbf{\texttt{input0}} & \textbf{\texttt{result0}} & \textbf{\texttt{str0}} & \textbf{\texttt{list0}} \\
\midrule
1  &    2 &  \textquotesingle ProgSys\textquotesingle &       0 &    \textquotesingle\textquotesingle   &                                 [] \\
2  &    4 &  \textquotesingle ProgSys\textquotesingle &       0 &    \textquotesingle P\textquotesingle &                               [80] \\
3  &    4 &  \textquotesingle ProgSys\textquotesingle &       0 &    \textquotesingle r\textquotesingle &                          [80, 114] \\
4  &    4 &  \textquotesingle ProgSys\textquotesingle &       0 &    \textquotesingle o\textquotesingle &                     [80, 114, 111] \\
5  &    4 &  \textquotesingle ProgSys\textquotesingle &       0 &    \textquotesingle g\textquotesingle &                [80, 114, 111, 103] \\
6  &    4 &  \textquotesingle ProgSys\textquotesingle &       0 &    \textquotesingle S\textquotesingle &            [80, 114, 111, 103, 83] \\
7  &    4 &  \textquotesingle ProgSys\textquotesingle &       0 &    \textquotesingle y\textquotesingle &       [80, 114, 111, 103, 83, 121] \\
8  &    4 &  \textquotesingle ProgSys\textquotesingle &       0 &    \textquotesingle s\textquotesingle &  [80, 114, 111, 103, 83, 121, 115] \\
9  &    5 &  \textquotesingle ProgSys\textquotesingle &     727 &    \textquotesingle s\textquotesingle &  [80, 114, 111, 103, 83, 121, 115] \\
10 &    6 &  \textquotesingle ProgSys\textquotesingle &      23 &    \textquotesingle s\textquotesingle &  [80, 114, 111, 103, 83, 121, 115] \\
11 &    7 &  \textquotesingle ProgSys\textquotesingle &      55 &    \textquotesingle s\textquotesingle &  [80, 114, 111, 103, 83, 121, 115] \\
12 &    8 &  \textquotesingle ProgSys\textquotesingle &       \textquotesingle 7\textquotesingle &    \textquotesingle s\textquotesingle &  [80, 114, 111, 103, 83, 121, 115] \\
\bottomrule
\end{tabularx}
\end{table*}

For these two benchmark problems, we see a huge difference in performance between the stack-based and the grammar-guided GP approaches which influences the high
variance in the results. For the Mirror Image problem, stack-based GP achieves a median success rate of nearly 100 while the grammar-guided GP approaches fail most
of the time. The results are similar for Vector Average, where stack-based GP has a median success rate of around 70 and grammar-guided GP achieves only a median
success rate lower than 10. Certainly we cannot conclude from these two results that stack-based GP generally performs better than grammar-guided GP, as grammar-guided GP, 
e.g., performs best on the Pig Latin problem (see Table \ref{tab:successful_sol}). Nevertheless, it shows that a comparison of different methods is still challenging despite 
all the efforts made in the benchmark suite as small differences in program synthesis approaches may lead to huge differences in performance. 
For example, program synthesis approaches often considerably differ with regard to the supported functions and structures which is problematic as the 
choice of the right functions has a huge impact on the problem solving performance. Forstenlechner et al. \cite{forstenlechner2018extending} already made the point that 
by adding specialized functions, a benchmark problem may become quite easy. For example, adding Python's \texttt{reverse()}, \texttt{round()}, and \texttt{sum()} functions 
to a grammar would turn Mirror Image and Vector Average into simple problems. However, the quality of a program synthesis approach that finds working solutions without such
specialized functions should be rated higher, as such an approach must be able to construct complex structures such as loops and conditionals which is more valuable in
practice. For future research, we recommend not only to provide the used grammar (or function set), but also making sure that the range of the used functions is similar when
comparing different methods. In addition, we recommend not to search explicitly for specialized functions which make a solution for the benchmark problems simple, but to 
strive for more general program synthesis methods. The recently published expansion of the benchmark suite \cite{helmuth2021psb2} may also help to achieve this goal, 
as we expect that a larger number of benchmark problems will direct the focus of program synthesis research towards generalizing approaches that work on many different problem types.

\subsection{In-Depth Analysis of a Complex Benchmark Problem}

To illustrate the main challenges in program synthesis with evolutionary algorithms in detail, we consider a complex problem from the benchmark suite as example. 
Figure \ref{fig:code_checksum} shows an exemplary hand-written Python function implementing a solution for the Checksum problem where the majority of the studied approaches fail to find 
a successful solution (see Table \ref{tab:exact_success_rates}). The given example code consumes more LOC than necessary but this allows us to analyze the code
line by line. The Checksum problem consists of several steps. For a successful solution, a given input string (\texttt{input0}, line 1) is converted to a list 
containing the ASCII value for every character of the input string (line 3-4). Then, the list is summed (line 5) and the resulting sum is taken modulo 64 (line 6).
Finally, the ASCII value of the space character is added (line 7) and then the temporary value is converted back to a character (line 8) before the result
is returned (\texttt{result0}, line 9) \cite{helmuth2015general}. 

\begin{figure}[!ht]
\begin{lstlisting}[frame=lines,language=Python,numbers=left,xleftmargin=2.4em,framexleftmargin=2.3em,basicstyle=\small]
def checksum(input0):
    result0, str0, list0 = 0, '', []
    for str0 in input0:
        list0.append(ord(str0))
    result0 = sum(list0)
    result0 %= 64
    result0 += ord(' ')
    result0 = chr(result0)
    return result0
\end{lstlisting}
\caption{Reference implementation in Python for the Checksum problem. \label{fig:reference_impl_checksum}}
\label{fig:code_checksum}
\end{figure}

It is not surprising that this program synthesis problem is quite challenging for evolutionary algorithms as it consists of many individual sub-problems. For a given set 
of input/output examples, we would not even expect to get a correct solution from an experienced human programmer as without additional information it is nearly impossible 
to identify the process required to map the input to a correct output.

In order to illustrate the program flow for a specific example input, Table~\ref{tab:program_trace} presents a program trace showing all variable states of the 
reference implementation for the Checksum problem (Fig.~\ref{fig:code_checksum}) for the input \textquotesingle ProgSys\textquotesingle. We see that several differently 
typed variables have to match exactly such that a correct result can be returned. Additionally, it is noticeable that it is not possible to assess the quality of the 
program based on the variable states in the first eleven execution steps. In the first eight execution steps the variable \texttt{result0 = 0} which gives us no information
about already correctly solved sub-problems. Also the integer values in steps 9-11 for \texttt{input0} give us no information as we only have input/output
examples (where the known output is a character) for training and during the evolutionary search we cannot see what happens inside the program. 
This is problematic as sub-problems that have already been solved correctly may be lost again during a continued search. 

However, with the current evolutionary search strategies a guided search for solutions to problems like the Checksum problem is not possible as they belong to the class 
of needle-in-a-haystack problems where finding a correct solution is strongly influenced by chance \cite{sobania2020challenges}. Consequently, we recommend
to focus research on uncovering these hidden sub-problems such that solutions can be generated step by step during evolution.
A software development method that may be suitable for a combination with evolutionary algorithms is test-driven development \cite{beck2003test}, where 
the software tests and the functional code are developed in parallel and large problems are regularly divided into several smaller sub-problems. 
Using this approach, program synthesis could be a real support for programmers in practice, since only the tests for the sub-problems have to be provided.

\section{Conclusions}\label{sec:conclusions}

The automatic generation of computer programs is one of the applications with practical relevance in evolutionary computation and especially in the field of GP. With 
program synthesis techniques programmers could be supported in everyday software development and also users without any knowledge in programming could easily automate their
repetitive tasks. 

To ensure comparability while studying the performance of novel evolutionary program synthesis approaches, in recent years usually the general program 
synthesis benchmark suite by Helmuth and Spector \cite{helmuth2015general} is used. The benchmark suite consists of 29 curated introductory programming 
problems of different complexity and also defines the structure of the training and test data. 

Using the program synthesis benchmark suite as starting point, we identified in this work the relevant approaches for program synthesis with evolutionary algorithms
and provided an in-depth analysis of the performance of the recent approaches on the benchmark problems.

As most relevant evolutionary approaches for program synthesis, we identified and described stack-based GP, grammar-guided GP as well as linear GP. 
Mainly, these three approaches differ in their way of representing the solutions, which results in individually different challenges and opportunities like, e.g., 
genotype-phenotype mapping or novel variation operators, respectively. However, for all three approaches, we see that in future research it will be necessary to generate
programs that are structurally similar to programs written by human programmers such that the resulting programs can be used in real-world software.

Further, we analyzed the performance of the recent program synthesis approaches on the individual benchmark problems reported in the literature. We found, that the 
approaches perform well on problems where a correct solution is small or if a program's input maps directly to its output. Problems are difficult for the recent
program synthesis approaches if they consist of several sub-problems, need type changes, or require iterative/recursive structures which interfere the mapping 
process from the inputs to the outputs. In such cases, there exists no strong fitness signal that guides the evolutionary search towards a correct solution. 
To overcome these problems, we encourage researchers to study methods that consider and assess also what happens inside a program such that useful solutions for sub-problems
or working loops and conditionals are also rewarded instead of only evaluating a program's output.

\section*{Acknowledgments}

We thank Marcel Hauck for the helpful discussion about large-scale language models and for providing access to the GPT-3 playground.

\bibliographystyle{IEEEtran}
\bibliography{IEEEabrv,main}

\end{document}